\documentclass[11pt, letterpaper]{arxiv}
\usepackage[all]{hypcap}
\usepackage[comma,numbers,sort,compress]{natbib}
\usepackage{hyperref}[citecolor=magenta]
\usepackage[capitalize,noabbrev]{cleveref}
\hypersetup{
    colorlinks = true,
    citecolor = {magenta},
}

\usepackage{microtype}

\usepackage{subcaption}
\usepackage{booktabs} %

\usepackage{algorithm}
\usepackage{algorithmic}
\usepackage{mathrsfs}
\usepackage{setspace}

\usepackage{amsmath}
\usepackage{amssymb}
\usepackage{mathtools}
\usepackage{amsthm}
\usepackage[inkscapelatex=false]{svg}

\setboolean{logo}{true}

\usepackage[most,skins,theorems]{tcolorbox}
\tcbset{
  aibox/.style={
    width=\linewidth,
    top=7pt,
    bottom=2pt,
    colback=blue!6!white,
    colframe=black,
    colbacktitle=black,
    enhanced,
    center,
    attach boxed title to top left={yshift=-0.1in,xshift=0.15in},
    boxed title style={boxrule=0pt,colframe=white,},
  }
}
\newtcolorbox{AIbox}[2][]{aibox,title=#2,#1}

\usepackage{wrapfig}
\definecolor{lightblue}{rgb}{0.22,0.45,0.70}%
\usepackage{tabularx}

\definecolor{rliableolive}{HTML}{BBCC33}
\definecolor{rliableblue}{HTML}{77AADD}
\definecolor{rliablered}{HTML}{EE8866}

\usepackage{crossreftools}
\usepackage[suppress]{color-edits}

\usepackage{dsfont}
\usepackage{nicefrac}
\newtcolorbox{analysisbox}[1][]{
    enhanced jigsaw,
    colback=white,
    colframe=blue!75!black,
    fonttitle=\bfseries,
    boxsep=5pt,
    left=5pt,
    right=5pt,
    top=5pt,
    bottom=5pt,
    title=#1,
}
\definecolor{editInitialResponse}{RGB}{255, 235, 156} %
\definecolor{editBacktrack}{RGB}{0, 0, 139} %
\definecolor{editRevisedResponse}{RGB}{255, 182, 193} %

\usepackage{pifont}
\usepackage{soul}
\definecolor{highlightmistake}{RGB}{255, 179, 179} 
\definecolor{highlightcorrect}{RGB}{179, 255, 179}

\usepackage[capitalize,noabbrev]{cleveref}

\theoremstyle{plain}
\newtheorem{theorem}{Theorem}[section]

\theoremstyle{definition}

\newtheorem{definition}[theorem]{Definition}
\theoremstyle{remark}

\newtcolorbox{solutionbox}{
  colframe=black,
  colback=gray!10,
  boxrule=1pt,
  arc=0pt,
  title=,
  fonttitle=\bfseries
}
\newenvironment{sol}
  {\begin{solutionbox}}
  {\end{solutionbox}}
\newcommand{\BeginSol}{\begin{sol}}
\newcommand{\EndSol}{\end{sol}}

\usepackage[textsize=tiny]{todonotes}

\makeatletter
\renewcommand\AB@authnote[1]{} % Removes the superscript numbers
\renewcommand\AB@affilnote[1]{} % Removes superscripts from affiliations
\makeatother
\usepackage{enumitem}

\usepackage{amsmath,amsfonts,bm}

\def\eqref#1{Eq.~\ref{#1}}

\def\1{\bm{1}}

\DeclareMathAlphabet{\mathsfit}{\encodingdefault}{\sfdefault}{m}{sl}
\SetMathAlphabet{\mathsfit}{bold}{\encodingdefault}{\sfdefault}{bx}{n}

\definecolor{codegray}{gray}{0.9}
\definecolor{codepurple}{rgb}{0.58,0,0.82}
\definecolor{codeblue}{rgb}{0.25,0.5,0.5}

\lstdefinelanguage{YAML}{
  morekeywords={selector, sequence, condition, task, no},   % Keywords to highlight (DSL specific)
  keywordstyle=\color{codeblue}\bfseries,                % Style for keywords
  ndkeywords={},                                        % No negative keywords
  sensitive=false,                                       % Case insensitive keywords
  comment=[l]{\#},                                      % Line comments start with #
  morecomment=[s]{/*}{*/},                               % Block comments (if any in your DSL, adjust if needed)
  commentstyle=\color{dkgreen}\ttfamily,               % Style for comments
  string=[b]",                                          % Strings in double quotes
  stringstyle=\color{codepurple}\ttfamily,              % Style for strings
  morestring=[b]',                                         % Strings in single quotes (if needed in your DSL)
  morestring=[b]`,                                         % Strings in backticks (if needed)
  identifierstyle=\ttfamily,                             % Style for identifiers (names)
  backgroundcolor=\color{codegray},                      % Background color for code block
  basicstyle=\ttfamily\footnotesize,
  breaklines=true,                                      % Break long lines
  captionpos=b,                                         % Caption position: bottom
  frame=single,                                        % Single line frame around listing
  numbers=left,                                        % Line numbers on the left
  numberstyle=\tiny\color{gray},                        % Style for line numbers
  numbersep=5pt,
  tabsize=2,                                           % Tab size
  showspaces=false,                                      % Don't show spaces as visible characters
  showstringspaces=false,                                % Don't show spaces in strings
  showtabs=false,                                        % Don't show tabs as visible characters
  xleftmargin=1em,
}
\title{Agents Play Thousands of 3D Video Games}

\author[]{Zhongwen Xu}
\author[]{Xianliang Wang}
\author[]{Siyi Li}
\author[]{Tao Yu}
\author[]{Liang Wang}
\author[]{Qiang Fu}
\author[]{Wei Yang}

\affil{\color{TencentBlue}{Tencent}}

\begin{document}
\maketitle

\textbf{Abstract:} 
We present PORTAL\footnote{Portal symbolizes a gateway that provides access to thousands of game worlds and it represents connection and transition between different gaming domains.}, a novel framework for developing artificial intelligence agents capable of playing thousands of 3D video games through language-guided policy generation. By transforming decision-making problems into language modeling tasks, our approach leverages large language models (LLMs) to generate behavior trees represented in domain-specific language (DSL). This method eliminates the computational burden associated with traditional reinforcement learning approaches while preserving strategic depth and rapid adaptability.
Our framework introduces a hybrid policy structure that combines rule-based nodes with neural network components, enabling both high-level strategic reasoning and precise low-level control. A dual-feedback mechanism incorporating quantitative game metrics and vision-language model analysis facilitates iterative policy improvement at both tactical and strategic levels. The resulting policies are instantaneously deployable, human-interpretable, and capable of generalizing across diverse gaming environments.
Experimental results demonstrate PORTAL's effectiveness across thousands of first-person shooter (FPS) games, showcasing significant improvements in development efficiency, policy generalization, and behavior diversity compared to traditional approaches. PORTAL represents a significant advancement in game AI development, offering a practical solution for creating sophisticated agents that can operate across thousands of commercial video games with minimal development overhead. Experiment results on the 3D video games are best viewed on our \href{https://zhongwen.one/projects/PORTAL}{project website}.

\begin{flushright}
\textit{An agent must learn from its experience. \\ -- Rich Sutton}
\end{flushright}

\begin{section}{Introduction}

Games have served as foundational testing grounds for artificial intelligence development since the field's inception~\cite{shannon1950programming,DQN,AlphaGo,AlphaStar,Dota2,Voyager}. They provide ideal environments for measuring and advancing cognitive capabilities including pattern recognition, memory utilization, reasoning, sophisticated planning and generalization. The pursuit of creating AI systems capable of generalizing effectively across diverse game environments represents a significant and ongoing challenge at the intersection of academic AI research and the commercial game industry.

The gaming landscape underwent a profound transformation with the rise of user-generated content (UGC) platforms, most notably pioneered by Roblox. UGC platforms have demonstrated unprecedented scale in both creation and consumption. By allowing users to design, build, and share their own games within a unified ecosystem, UGC games fundamentally altered the game development paradigm.

The explosive growth of UGC gaming created a unique challenge for traditional game-playing AI systems. While conventional AI approaches had made remarkable progress in mastering specific games with well-defined rulesets -- from Go~\cite{AlphaGo} to StarCraft II~\cite{AlphaStar} -- they faced inherent limitations in environments characterized by constant flux and unbounded creativity. Traditional game AI typically relies on extensive domain-specific engineering, carefully crafted heuristics, or reinforcement learning methodologies that require millions/billions of training iterations within a single game environment. Such approaches simply cannot scale to match the pace and diversity of user-created content, where thousands of new games with unique mechanics, aesthetics, and objectives emerge daily. This fundamental mismatch between traditional AI capabilities and the rapidly expanding universe of user-generated games highlights a critical need for more adaptive, generalizable AI approaches -- a gap that we are aiming to bridge here.

This paper introduces PORTAL(\textbf{P}olicy \textbf{O}ptimization and \textbf{R}easoning for \textbf{T}actical \textbf{A}rtificial \textbf{L}earning), a novel approach to generating game-playing artificial intelligence (AI) agents capable of operating across thousands of 3D video games. By leveraging large language models (LLMs)~\cite{GPT4,claude3_7,Gemini,qwen25,qwen25coder} to generate specialized behavior trees (BTs)~\cite{BT} expressed in domain-specific language (DSL), we establish a new paradigm for AI agent development that combines the strategic reasoning of LLMs with the real-time performance requirements of commercial video games.
Unlike traditional reinforcement learning approaches~\cite{AlphaStar,Dota2,wzry,IMPALA,PPO} that require massive simulation resources and distributed training, our method decouples the tactical planning process from execution, enabling rapid development and deployment of sophisticated agents. This decoupling allows for the generation of diverse, adaptable, and human-like behaviors while maintaining the computational efficiency necessary for commercial application.
PORTAL represents a significant advancement in the field by transforming complex decision-making problems into language modeling challenges. Through an innovative hybrid architecture that integrates rule-based nodes with neural network components, we demonstrate how LLM-generated policies can be effectively executed in dynamic, real-time environments without the latency limitations that typically constrain LLM-based agents.

On methodology, we propose a \emph{meta algorithm} for decision-making problems that fundamentally reimagines the role of large language models (LLMs) in game AI systems. Unlike prior works~\cite{Voyager,llm_sc2,claude3_7} where LLMs directly function as \emph{actor policies}, our approach positions these models as \emph{architects} that generate \emph{policy structures}. This important innovation creates a hybrid architecture that leverages the strengths of both paradigms: LLMs design sophisticated policy frameworks composed of interconnected basic building blocks, with each node representing discrete functionality. These nodes can then be implemented as neural networks, enabling dynamic adjustment to changing game environments. The nodes can also be implemented with rule-based codes, where it brings exact control over behaviors. The resulting system maintains the strategic sophistication and broadly contextual awareness of LLMs while incorporating the adaptability and responsiveness of neural network approaches.

The remainder of this paper is organized as follows: Section~\ref{sec:related} reviews related work in behavior trees, reinforcement learning, and language models for game playing AIs. Section~\ref{sec:method} details our methodology, including the hybrid policy structure, DSL representation, and an agent-environment loop with reflection feedbacks. Section~\ref{sec:exps} presents experimental results, and Section~\ref{sec:dis} discusses implications and potential applications beyond gaming. We conclude our work and point out directions for future research.
\end{section}
\begin{section}{Related Work}\label{sec:related}

\textbf{Behavior Trees}: Behavior Trees (BTs)~\cite{BT} have established themselves as a predominant paradigm for artificial intelligence in modern game development, particularly within sophisticated engines such as Unreal Engine 5\footnote{\href{https://dev.epicgames.com/documentation/en-us/unreal-engine/unreal-engine-behavior-tree-node-reference-tasks}{Unreal Engine 5 Documentation on Behavior Trees}} and Unity. Their popularity stems from their capacity to organize complex AI behaviors in a structured, modular, and visually intuitive manner. The foundational components of BTs include:

\begin{itemize}
    \item \texttt{Selector} Nodes: Implement prioritized decision-making by evaluating child nodes sequentially until one succeeds, enabling fallback behaviors.
    \item \texttt{Sequence} Nodes: Execute a series of actions in order, succeeding only if all component actions succeed, facilitating complex, multi-step behaviors.
    \item \texttt{Task} Nodes: Perform atomic game actions such as movement, attack sequences, or environmental interactions.
    \item \texttt{Condition} Nodes: Evaluate environmental states to determine execution paths.
\end{itemize}

This architecture is complemented by a Blackboard system that maintains state information accessible to all nodes within the tree. In commercial implementations such as Unreal Engine 5, a dedicated AI controller periodically evaluates the behavior tree against current game state, determining appropriate actions for non-player characters (NPCs). 

The modular structure of BTs facilitates code reuse, simplifies debugging, and enables designers to implement sophisticated behaviors without extensive programming expertise. However, traditional BTs rely primarily on manually crafted rules and conditions, limiting their adaptability to novel or highly dynamic scenarios. Our work extends this established framework by incorporating neural components and LLM-based generation, preserving the interpretability and modularity of classical BTs while enhancing their adaptability and generalization capabilities.

\textbf{Reinforcement Learning}: Reinforcement Learning (RL)~\cite{RL_book} has demonstrated remarkable achievements in mastering complex environments across various domains. AlphaGo~\cite{AlphaGo}'s victory over world champion Lee Sedol in 2016 marked a watershed moment, demonstrating how deep neural networks combined with Monte Carlo Tree Search (MCTS) could achieve superhuman performance in challenging strategic domains. This approach initially leveraged human expert data before transitioning to pure self-play in subsequent iterations such as AlphaGo Zero~\cite{AlphaGoZero} and AlphaZero~\cite{AlphaZero}, which surpassed its predecessor's capabilities without human guidance. Beyond board games, RL has tackled increasingly complex video game environments. OpenAI Five~\cite{Dota2} demonstrated emergent team strategies in Dota 2, while DeepMind's AlphaStar~\cite{AlphaStar} achieved Grandmaster-level performance in StarCraft II -- both representing significant advances in handling high-dimensional state and action spaces with partial observability. Similarly, Tencent's Honor of Kings AI system~\cite{wzry} has pushed the boundaries further in complex Multiplayer Online Battle Arena (MOBA) games.

These systems~\cite{PPO,IMPALA} typically employ large-scale distributed architectures, training on millions or billions of game frames to develop effective policies. While impressive in their capabilities, these approaches incur substantial computational costs -- often requiring hundreds or thousands of GPUs and enormous CPUs for game simulations, operating for weeks or months -- and frequently struggle with generalization beyond their specific training environments. Our work addresses these limitations by decoupling strategic planning from execution, significantly reducing computational requirements while enhancing cross-environment generalization.

\textbf{Language Models for Game-playing AIs}: Recent research has begun exploring the potential of large language models (LLMs) for creating game-playing agents. As comprehensively documented in the survey by~\cite{llm_games}, these approaches have primarily focused on three categories of games:

\begin{itemize}
    \item  \textbf{Abstract Token Games}: Board games like Chess~\cite{AlphaZero} and Go, where states and actions can be represented as discrete, symbolic tokens;
    \item  \textbf{Text-Based Games}: Interactive fiction environments like Zork~\cite{Zork}, where game interactions naturally occur through textual commands and descriptions;
    \item  \textbf{API-Mediated Games}: Complex environments where game state is transformed into textual descriptions and LLMs generate actions as API calls to game interfaces.

\end{itemize}

Voyager~\cite{Voyager} exemplifies the API-mediated approach, leveraging GPT-4~\cite{GPT4} to generate code for Minecraft tasks through the Mineflayer~\cite{mineflayer} JavaScript API. A distinctive feature of Voyager is its curriculum-based skill acquisition, progressively building a library of capabilities through tool usage and exploration. However, Voyager fundamentally relies on textual representations of game states rather than direct engagement with raw environmental data. Other research~\cite{llm_sc2} has applied similar principles to complex strategy games like StarCraft II through frameworks like TextStarCraft, which translates game states into textual descriptions and facilitates LLM interaction through API calls. While demonstrating the potential of LLMs for strategic reasoning, these approaches face a significant limitation: the inherent latency of LLM inference renders them impractical for real-time applications, with reported gameplay durations extending to seven hours per match -- clearly prohibitive for commercial deployment.
More recently, Anthropic's Claude 3.7 Sonnet~\cite{claude3_7} has demonstrated impressive capabilities in Pokémon Red with zero-shot generalization, defeating Gym Leaders through extended reasoning. However, its success relies on relaxed time constraints that would be unacceptable in commercial gaming contexts.
Our approach differs fundamentally from these precedents in three critical aspects:

\begin{itemize}
    \item \textbf{Offline Policy Generation}: Rather than relying on LLM inference during gameplay, we utilize LLMs to generate policies expressed in Domain-Specific Language (DSL), which are then interpreted into efficient, executable code, eliminating inference latency during gameplay.

    \item \textbf{Hybrid Architecture}: We combine the strategic reasoning capabilities of LLMs with the real-time performance of neural networks and rule-based systems, leveraging the strengths of each approach while mitigating their individual limitations.

    \item \textbf{Cross-Game Generalization}: To our knowledge, we are the first to demonstrate agents capable of playing thousands of distinct 3D video games through a unified approach, highlighting the generalization capabilities of our method.

\end{itemize}

This synthesis of LLM-based policy generation with efficient execution mechanisms represents a novel contribution to the field, addressing the practical constraints that have limited the application of language models in real-time gaming environments.

\end{section}
\section{Method}\label{sec:method}
\subsection{Overview}
Our approach introduces a novel policy representation that formalizes the hybrid structure of game-playing AI agents. We define a policy $\pi$ as a triple $(\Pi, \Theta, \Phi)$, where:

\begin{itemize}
    \item $\Pi$ represents the hierarchical tree structure that defines the control flow and relationships between nodes. Formally, $\Pi$ can be expressed as a directed acyclic graph (DAG) where nodes are drawn from $\Theta \cup \Phi$, and edges represent the execution flow between components;
    \item $\Theta$ denotes the set of neural network-parameterized task nodes: $\Theta = \{\theta_1, \theta_2, \dots, \theta_m\}$. Each $\theta_i$ implements a mapping from an observation space $O$ (a subset of the complete environment state $S$) to either a specific action or a probability distribution over possible actions. Mathematically, this can be expressed as $\theta_i: O \rightarrow A$ or $\theta_i: O \rightarrow P(A)$, where $A$ represents the action space;
    
    \item $\Phi$ comprises the set of rule-based nodes: $\Phi = \{\phi_1, \phi_2, \dots, \phi_n\}$. These include condition nodes and simple action nodes implemented through traditional programming. Each $\phi_j$ represents a deterministic function that maps an observation to either a boolean value (for condition nodes) or a specific action (for action nodes). This relationship can be formalized as $\phi_j: O \rightarrow \{\text{True}, \text{False}\}$ or $\phi_j: O\rightarrow A$.
\end{itemize}

This formalization offers several advantages over conventional approaches. First, it establishes a clear separation between the strategic structure of the policy ($\Pi$) and its implementation details ($\Theta$ and $\Phi$). This separation enables independent optimization of each component, facilitating rapid iteration and refinement. Second, by combining rule-based and neural network components, the policy can leverage both the interpretability and reliability of hand-crafted rules as well as the adaptability and pattern recognition capabilities of neural networks. Finally, this representation provides a natural framework for language models to generate and modify policies by focusing on the structural aspects ($\Pi$) while leaving the implementation details of neural components ($\Theta$) to specialized training procedures.

Our policy representation can be formally defined as:

\begin{tcolorbox}[colback=green!5!white,colframe=black,boxsep=0pt,top=4pt,bottom=4pt,left=3pt,right=3pt]
\begin{definition}
A policy $\pi = (\Pi, \Theta, \Phi)$ consists of a tree structure $\Pi$, neural network task nodes $\Theta = \{\theta_1, \theta_2, \dots, \theta_m\}$, and rule-based nodes $\Phi = \{\phi_1, \phi_2, \dots, \phi_n\}$, where $\Pi$ determines the hierarchical organization and execution flow of elements from $\Theta$ and $\Phi$.
\end{definition}
\end{tcolorbox}

\subsection{Domain-specific Language Representation}
We have developed a domain-specific language (DSL) to represent behavior trees in a format that is both human-readable and machine-executable. This DSL serves as the interface between large language models and the policy execution environment, enabling seamless translation of strategic concepts into operational policies.

The syntax of our DSL follows a hierarchical structure that directly mirrors the organization of behavior trees. Key syntactic elements include:

\begin{itemize}
    \item \textbf{Hierarchical Structure}: Indentation denotes parent-child relationships between nodes, providing a visual representation of the tree's organizational structure. This design choice enhances readability while maintaining a direct correspondence to the underlying computational representation.

    \item \textbf{Node Types}: The DSL supports four primary node types, each serving a distinct role in policy execution:

        \begin{itemize}
            \item         Selector Nodes (\texttt{selector:}): Implement prioritized execution, attempting child nodes sequentially until one succeeds.

            \item Sequence Nodes (\texttt{sequence:}): Execute child nodes in order, continuing only if each node succeeds.

            \item Condition Nodes (\texttt{condition:<condition\_key>}): Evaluate environmental predicates, determining execution paths based on game state.
    
            \item Task Nodes (\texttt{task: <action\_key><param\_key>}): Execute specific actions within the game environment, potentially utilizing neural networks for implementation.

            \item Logical Operations: The DSL incorporates logical operations, such as negation (\texttt{condition: no}), enabling the construction of complex conditional statements without requiring additional syntax.

        \end{itemize}

\end{itemize}

The following example illustrates a simple policy for a first-person shooter (FPS) game:

\begin{lstlisting}[language=YAML, caption=A Behavior Tree DSL Example]
selector:
  sequence:
    condition: has_enemy_in_view
    task: shoot random_enemy_in_view
  sequence:
    condition: no
    condition: has_enemy_in_view
    task: move_to random_enemy_location
\end{lstlisting}

This policy implements a straightforward combat strategy: if an enemy is visible, select a random visible enemy and attack; otherwise, move toward a known enemy location. The simplicity and clarity of this representation belies the sophisticated decision-making process it encodes, demonstrating the expressive power of our DSL.

A critical advantage of this representation is its compatibility with large language models, which can generate, interpret, and modify these tree structures based on high-level strategic descriptions. Furthermore, the DSL's hierarchical nature facilitates incremental refinement and composition of policies, enabling the construction of complex behaviors from simpler building blocks. This property is particularly valuable for the iterative improvement process employed in our approach.

\subsection{Extensions Beyond Classic Behavior Trees}
Traditional behavior trees typically rely exclusively on rule-based implementations, employing deterministic algorithms like A* pathfinding for navigation tasks. While effective in static environments, these conventional approaches falter when confronted with dynamic scenarios featuring moving obstacles, adversarial agents, or complex environmental interactions. We address this limitation by introducing a hybrid architecture that augments the classical behavior tree framework with neural network-parameterized task nodes.

\begin{figure}[htp]
\centering
\includegraphics[width=0.7\linewidth]{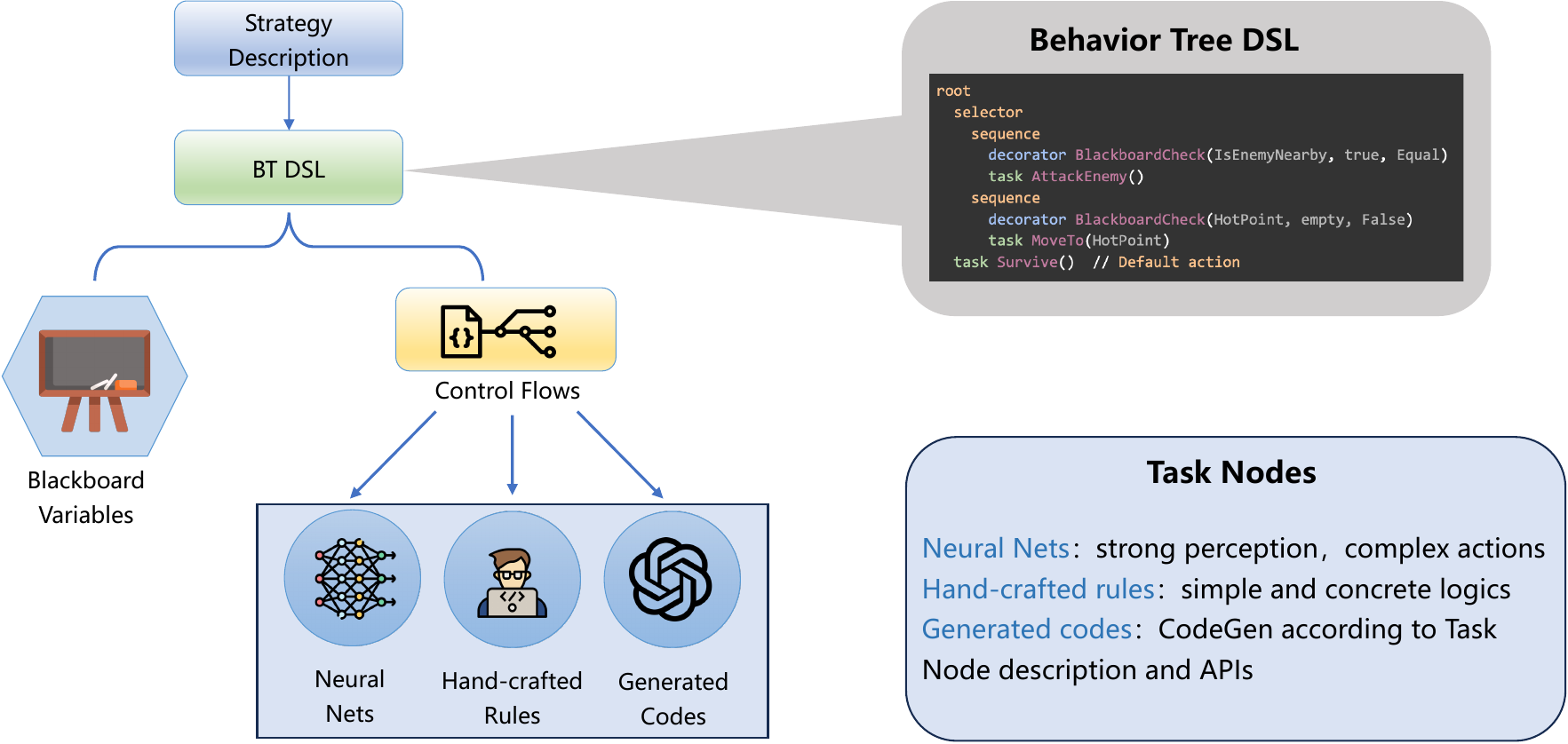}
\caption{Overview of the behavior tree generation process using LLMs.}
\label{fig:bt_gen}
\end{figure}

As shown in Figure~\ref{fig:bt_gen}, our architecture instantiates specific task nodes as neural networks, particularly for operations that demand adaptive responses to environmental complexity. For instance, rather than implementing a $\texttt{move\_to}$ task with a deterministic pathfinding algorithm, we utilize a neural network trained to navigate dynamic environments while avoiding obstacles and responding to changing conditions. \emph{Note that the neural nets applied in our work are tiny nets such as two layers of fully-connected layers / convolutional layers.}

This hybrid approach represents a fundamental advancement over both purely rule-based behavior trees and end-to-end neural policies. By combining the strengths of symbolic reasoning (through the behavior tree structure) with subsymbolic learning (via neural networks), we achieve policies that are simultaneously more adaptive, interpretable, and generalizable than either approach in isolation.

A fundamental innovation in our approach is the reconceptualization of decision-making problems as language modeling tasks. Rather than training policies through direct interaction with environments, we leverage the capabilities of LLMs to generate DSL representations of behavior trees. This transformation offers several compelling advantages:
\begin{itemize}
    \item \textbf{Rapid Iteration}: The language modeling approach facilitates rapid prototyping and refinement of policies, as modifications can be made at the structural level without requiring extensive retraining of neural components.
    \item \textbf{Enhanced Environmental Perception}: Neural network-parameterized task nodes leverage sophisticated perceptual capabilities to process environmental information, enabling more informed decision-making in complex scenarios. Rather than responding to simplified abstractions of the environment, these nodes can integrate rich sensory data to guide their actions.
    \item \textbf{Simplified Reward Structures}: Each neural network task node is designed with a focused, often unidimensional reward function. For example, a $\texttt{move\_to}$ node might optimize solely for minimizing distance to a target location, while a shoot node maximizes accuracy against a specific enemy. This decomposition of complex, multi-objective reward landscapes into simpler, targeted optimization problems significantly enhances learning efficiency and generalization capabilities.
    % \item \textbf{Domain Adaptation}: The separation of strategic decision-making (encoded in the behavior tree structure) from tactical execution (implemented by neural networks) facilitates rapid adaptation to new domains. The high-level strategy can remain constant while the underlying neural implementations are fine-tuned for specific environments, enabling efficient transfer learning.
\end{itemize}
The power of the proposed system demonstrates that even with tiny neural networks $\Theta$
, a well-designed Domain-Specific Language (DSL) can leverage them to construct a powerful policy $\pi$.
\begin{AIbox}{Transformation from Decision-Making into Language Modeling}
We transform the decision-making problems into language modeling problems of building up DSL for policies. This transformation allows us to leverage the full potential of LLMs as reasoning engines for strategic planning, while maintaining the computational efficiency necessary for real-time execution in dynamic environments.

\end{AIbox}

\subsection{An Agent-Environment Loop}
\begin{figure}[h!]
\centering
\includegraphics[width=0.6\linewidth]{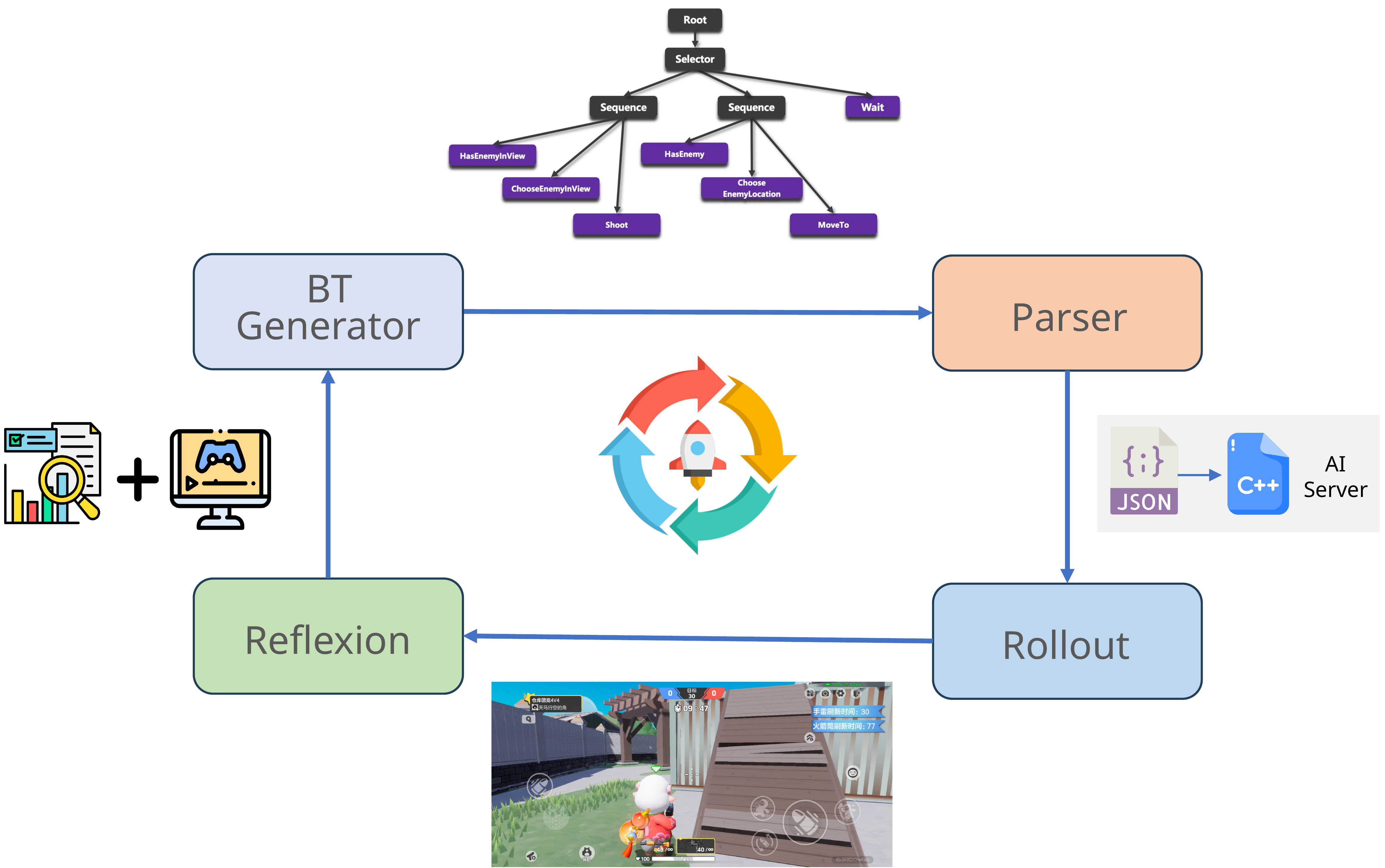}
\caption{System architecture showing the integration of LLM-based behavior tree generation with environment interaction and reflexion}
\label{fig:overview}
\end{figure}

Following the reinforcement learning paradigm, we conceptualize the interaction between our agents and their environments as a cyclical process of policy refinement. However, our approach diverges from traditional methods in its implementation of this cycle. The agent continuously refines its policy parametrization, denoted as $\Pi$, through structured interaction with the environment and analysis of the resulting outcomes.

Central to our approach is the Reflexion~\cite{reflexion, ReAct} module, which serves as the bridge between environmental feedback and policy improvement. This module processes two distinct types of feedback:

\begin{itemize}
    \item \textbf{Quantitative Game Metrics}: Direct numerical measurements extracted from the game environment, such as kills, deaths, objective completions, and resource utilization. These metrics provide precise, objective evaluations of policy performance across specific dimensions. To make these metrics accessible to language models, we translate them into natural language descriptions that highlight relevant patterns and trends.
    \item \textbf{Vision-Language Model Analysis}: A complementary feedback mechanism that utilizes a vision-language model (VLM) to analyze "mini-map" representations of gameplay from a bird's-eye perspective. This analysis provides high-level strategic insights that may not be captured by numerical metrics alone, such as spatial control patterns, tactical adaptations, and coordination effectiveness.

\end{itemize}

The dual-feedback mechanism enables comprehensive policy evaluation and improvement at multiple levels of abstraction. The quantitative metrics offer granular feedback on tactical execution, while the VLM analysis provides broader strategic context. By integrating these complementary perspectives, our system can identify both local optimizations and global strategic shifts that may improve overall performance.
The iterative improvement process follows a structured cycle: the LLM generates an initial behavior tree expressed in DSL, based on high-level strategic descriptions.
This behavior tree is compiled into an executable policy and deployed in the game environment.
The agent interacts with the environment, generating gameplay data.
The Reflexion module processes this data, extracting quantitative metrics and producing VLM analyses.
These insights are provided to the LLM, which generates an improved behavior tree that addresses identified weaknesses. This cycle continues iteratively, with each generation refining the policy based on accumulated insights from previous deployments. The result is a continuously improving agent that adapts to the specific challenges and opportunities presented by its environment. The overall procedure is shown in Figure~\ref{fig:overview}.

\begin{AIbox}{Policy Improvement with Experience}
Through Reflexion on environmental experience, the LLM updates the policy DSL $\Pi$ to optimize for either quantitative game metrics or global tactical analysis.
\end{AIbox}

\subsection{Chain-of-Thought for Tree Generation and Reflexion}
The hierarchical nature of our policy representation aligns naturally with a structured reasoning approach. Our policies exhibit a tree-like organization where higher levels manage strategic planning while lower levels execute tactical actions to achieve specific objectives. For instance, within this hierarchy, \texttt{Sequence} nodes decompose complex goals into sequential subtasks executed from left to right, while \texttt{Selector} nodes implement prioritized decision-making, attempting child nodes in order until one succeeds.

To leverage this hierarchical structure effectively, we implement a level-by-level generation process utilizing Chain-of-Thought (CoT) prompting~\cite{CoT}. This technique instructs the LLM to construct the behavior tree progressively, beginning with root nodes and systematically expanding through each subsequent level. This approach offers several advantages -- \textbf{Focused Reasoning}: By addressing one level at a time, the LLM can concentrate its reasoning capacity on a manageable subset of the policy structure, improving overall coherence and strategic alignment; \textbf{Iterative Refinement}: The level-by-level approach facilitates targeted revision of specific tree components without requiring complete regeneration of the entire policy.

Complementing this generative process, we implement a level-by-level reflexion mechanism that evaluates and revises specific sections of the tree based on performance feedback. When behavioral deficiencies are identified, the LLM can focus its analysis on the relevant subtree, proposing targeted modifications that address specific shortcomings while preserving effective components.

Our CoT implementation employs structured prompting with explicit reasoning steps, guiding the LLM through a systematic tree construction process: First, the LLM identifies the primary strategic objectives and potential challenges.
Next, it determines the appropriate root node type (typically a Selector) to organize high-level decision-making.
For each child of the root, the LLM articulates a specific subgoal and constructs an appropriate subtree.
This process continues recursively until all paths terminate in executable task nodes. Throughout this process, the LLM explicitly documents its reasoning, explaining why particular node types, conditions, and actions were selected. This explicitness not only improves the quality of the generated policies but also enhances interpretability, providing human designers with insight into the strategic principles underlying the policy structure.

\begin{AIbox}{Key Takeaways}
Tree-like CoTs facilitate better reasoning and reflection in strategy planning.
\end{AIbox}

\subsection{Sampling and Post-training}
\begin{figure}[htp]
\centering
\includegraphics[width=0.6\linewidth]{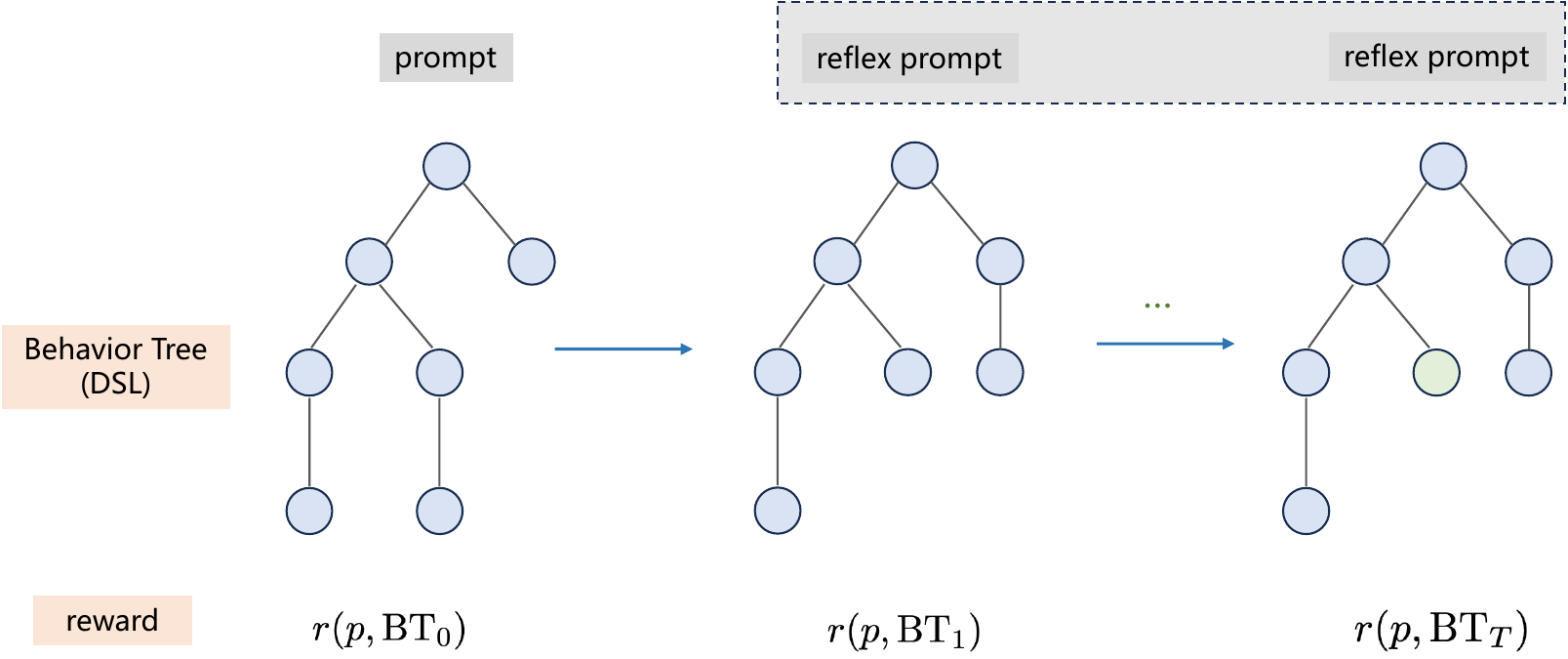}
\caption{The LLM-based behavior tree generation and refinement process}
\label{fig:llm_process}
\end{figure}

By recasting decision-making as a language modeling problem, we can generate extensive training data through systematic exploration of the policy space. This exploration utilizes breadth-first search (BFS) to sample $N$ potential behavior trees in parallel, retaining the $K$ best-performing policies in each iteration based on direct environmental rewards. 

The sampling process generates valuable trajectory data that can be leveraged for model improvement. Unlike traditional reinforcement learning, which collects environment trajectory data (i.e., state-action pairs), our approach collects LLM trajectory data -- sequences of prompts, DSL outputs, and associated rewards:
\begin{equation*}
    [(\text{prompt}_0, \text{DSL}_0, r_0),   (\text{prompt}_1, \text{DSL}_1, r_1),   (\text{prompt}_2, \text{DSL}_2, r_2),  \dots (\text{prompt}_T, \text{DSL}_T, r_T)]
\end{equation*}

This data can be utilized for Supervised Fine-Tuning (SFT) or reinforcement learning approaches tailored to language models. The structured format directly connects strategic descriptions, policy implementations, and performance outcomes, enabling the model to learn the relationships between high-level goals and effective policy structures.

As illustrated in Figure~\ref{fig:llm_process}, each iteration $i$ produces a specific DSL output representing a policy and receives reward feedback $r_i$ from the environment. By structuring each step in the LLM inference process with explicit tags like \texttt{<reflection>...</reflection>} and \texttt{<think>...</think>}~\cite{R1}, we create "Meta CoTs"~\cite{meta_cot} that enable the model to systematically explore the solution space and iteratively improve its policy outputs.

\subsection{A Policy Network to Schedule Policies}
The generative capabilities of LLMs, whether frozen or fine-tuned, enable the production of diverse policy candidates expressed as DSL descriptions. These candidates represent various strategic approaches to addressing game challenges. However, this diversity introduces a new challenge: determining which policy is most appropriate for a specific environment state at a given moment. While exhaustive evaluation of all candidates is theoretically possible, it would be computationally prohibitive in real-time applications.

To address this challenge, we introduce a policy scheduling network that dynamically selects from a repertoire of pre-generated behavior trees based on environmental conditions. This network receives the same state observations as traditional reinforcement learning policies but produces a fundamentally different output. Rather than directly selecting atomic actions, it identifies the most suitable behavior tree for the current context from a pre-defined library of options.

This approach draws inspiration from the options~\cite{option,RL_book} framework in hierarchical reinforcement learning, where options represent temporally extended courses of action that an agent can follow. In our implementation, each behavior tree constitutes an option -- a predetermined policy that, once selected, governs the agent's behavior until completion or interruption. The policy network operates at a meta-decision level, determining when to switch between these options based on changing environmental conditions.
The policy network receives environmental observations and selects the most appropriate behavior tree from the available library. This tree is then executed until completion or until the network determines that a different strategy would be more effective given updated environmental conditions.

\section{Experiments}\label{sec:exps}
We conducted experiments on Yuan Meng Star\footnote{\href{https://en.wikipedia.org/wiki/Yuan\_Meng\_Star}{https://en.wikipedia.org/wiki/Yuan\_Meng\_Star}}, a platform developed by TiMi Studio under Tencent Games, which hosts tens of millions of User Generated Content (UGC) games. Yuan Meng Star provides players with a comprehensive UGC platform featuring built-in editors, tools, and assets to create their own games. The platform encompasses a diverse range of game genres, including speedrunning, survival, first-person shooter (FPS), and party games. For our research, we demonstrated the efficacy of our proposed method specifically on FPS games created by thousands of players, though the same principles can be applied to develop agents for other game genres as well. The LLMs used to produce the DSL are \texttt{Qwen2.5-32B-Coder}~\cite{qwen25coder}. All the materials of the shown BTs can be found on our \href{https://zhongwen.one/projects/PORTAL}{project website}, including the illustration, DSL and JSON files of the BTs. The neural networks in the task nodes have only two Convolutional layers + Fully-connected layers. \emph{All the experiments shown in the following are with a frozen LLM without post-training.}

Due to copyright restrictions associated with UGC games, we present only a select subset of representative games in this paper. However, it should be noted that comprehensive testing across a substantially broader range of UGC environments has been conducted internally, with consistent results supporting our findings. The examples showcased here were specifically chosen to illustrate key aspects of our methodology while respecting intellectual property constraints.

\subsection{Optimization for Game Metrics}

\begin{figure}[h!] % Use [h!] to encourage placement here, but adjust as needed
\centering % Center the whole figure environment

\begin{subfigure}{0.48\textwidth} % Adjust width as needed
  \centering
  \includegraphics[width=\linewidth]{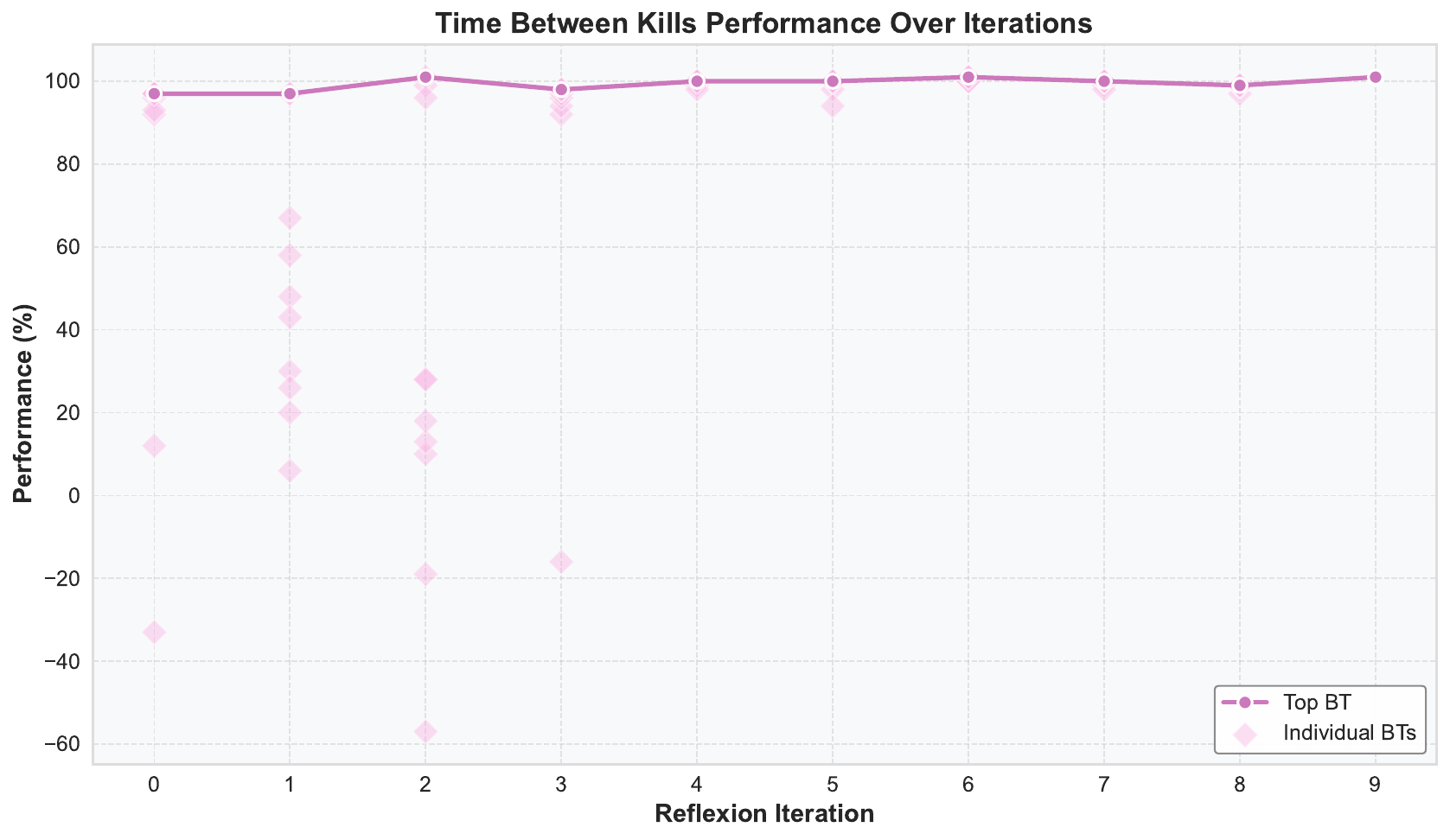} % Replace with your first image file
  %\caption{First subfigure}
  \label{fig:sub1}
\end{subfigure}\hfill % Key for horizontal spacing: \hfill pushes them apart
\begin{subfigure}{0.48\textwidth} % Adjust width as needed
  \centering
  \includegraphics[width=\linewidth]{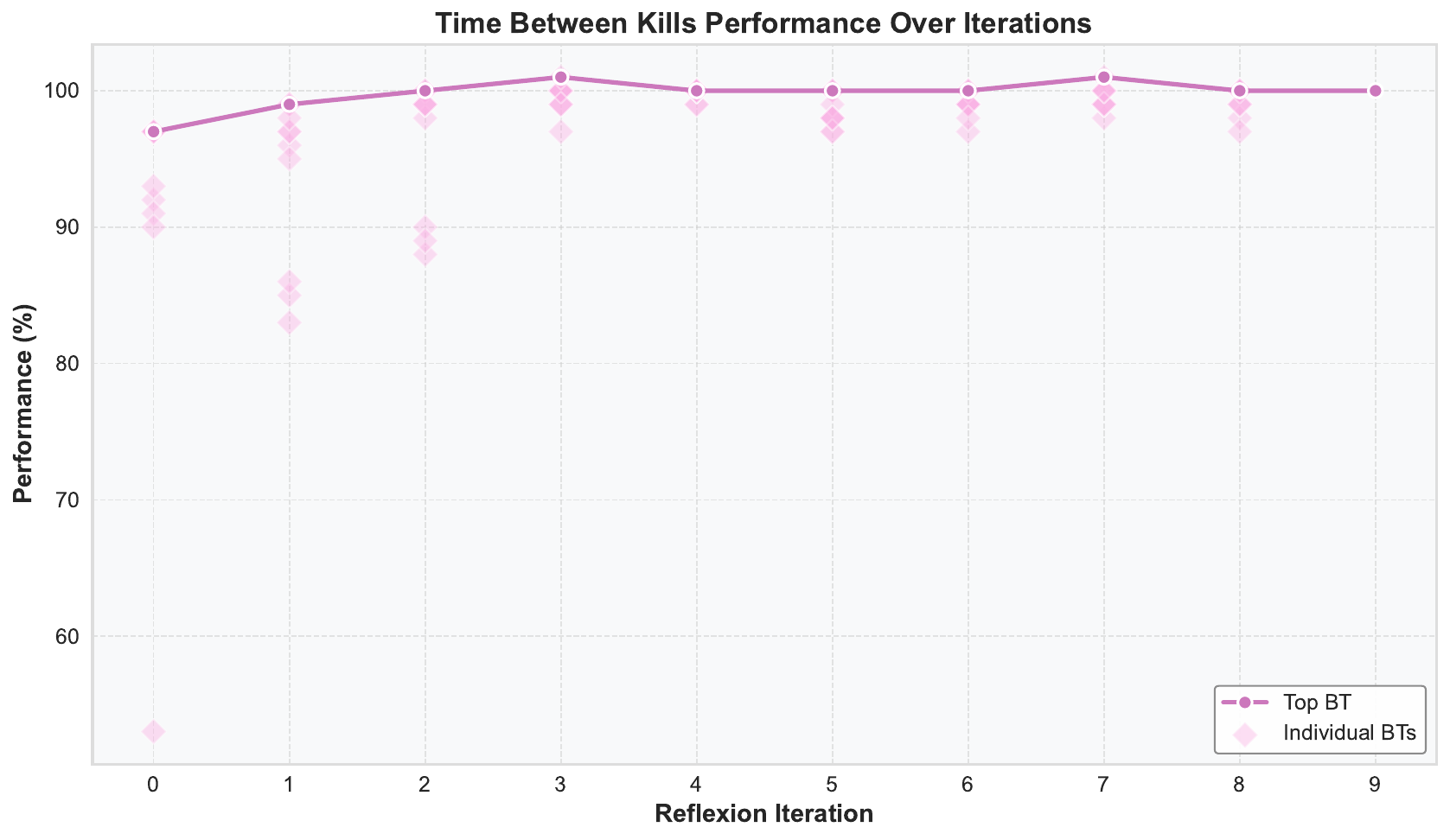} % Replace with your second image file
  %\caption{Second subfigure}
  \label{fig:sub2}
\end{subfigure}
\caption{Two independent runs of Breadth-first Search (BFS). }
\label{fig:metric_reflex}
\end{figure}

We first demonstrate the effectiveness of our approach through a Breadth-first Search (BFS) procedure designed to optimize specific game metrics. Figure~\ref{fig:metric_reflex} illustrates two independent experimental runs optimizing the "time between kills" metric, where 100\% represents optimal performance. Individual Behavior Trees (BTs) are depicted in lighter colors, while the top-performing BT is highlighted with solid dots. The results clearly show that through successive reflection iterations, the individual BTs converge toward near-optimal performance. Notably, the top-performing BTs maintain consistent performance across iterations. These findings confirm that our proposed algorithm, through structured agent-environment interactions, successfully optimizes specified game metrics by refining the underlying policy structure. This demonstrates the viability of our meta-algorithmic approach in which LLMs architect policies rather than directly outputting actions.

\subsection{Vision-Language Models for Reflexion}

Vision-Language Models (VLMs) such as Gemini~\cite{Gemini} provide powerful video understanding capabilities that can analyze extended gameplay footage through natural language interaction. This multimodal understanding capacity offers a valuable mechanism for tactical planning assessment and refinement. In traditional competitive games like Dota and StarCraft II, mini-maps provide players with critical global information to inform strategic decision-making. While the games in our experimental environment do not natively provide such mini-maps, we generate them synthetically using environmental data including map layouts, player positions, status indicators, and other relevant information.
We accumulate this information throughout gameplay sessions and feed the resulting replay recordings into a VLM, enabling comprehensive analysis of global tactical strategies. For our evaluation of first-person shooter games, we structure this analysis across five critical dimensions:

\begin{itemize}
    \item \textbf{Map Control}: Assessment of territorial dominance, strategic position maintenance, and spatial resource utilization.
    \item \textbf{Adaptability}: Evaluation of the agent's capacity to adjust strategies in response to changing circumstances or opponent behaviors.
    \item \textbf{Team Coordination}: Analysis of collaborative behaviors, role specialization, and synchronized actions.
    \item \textbf{Team Aggression}: Measurement of offensive initiative, pressure application, and risk-taking behavior.
    \item \textbf{Goal Achievement}: Assessment of progress toward primary and secondary objectives.
\end{itemize}

\begin{figure}[htp]
    \centering
    \includegraphics[width=0.5\linewidth]{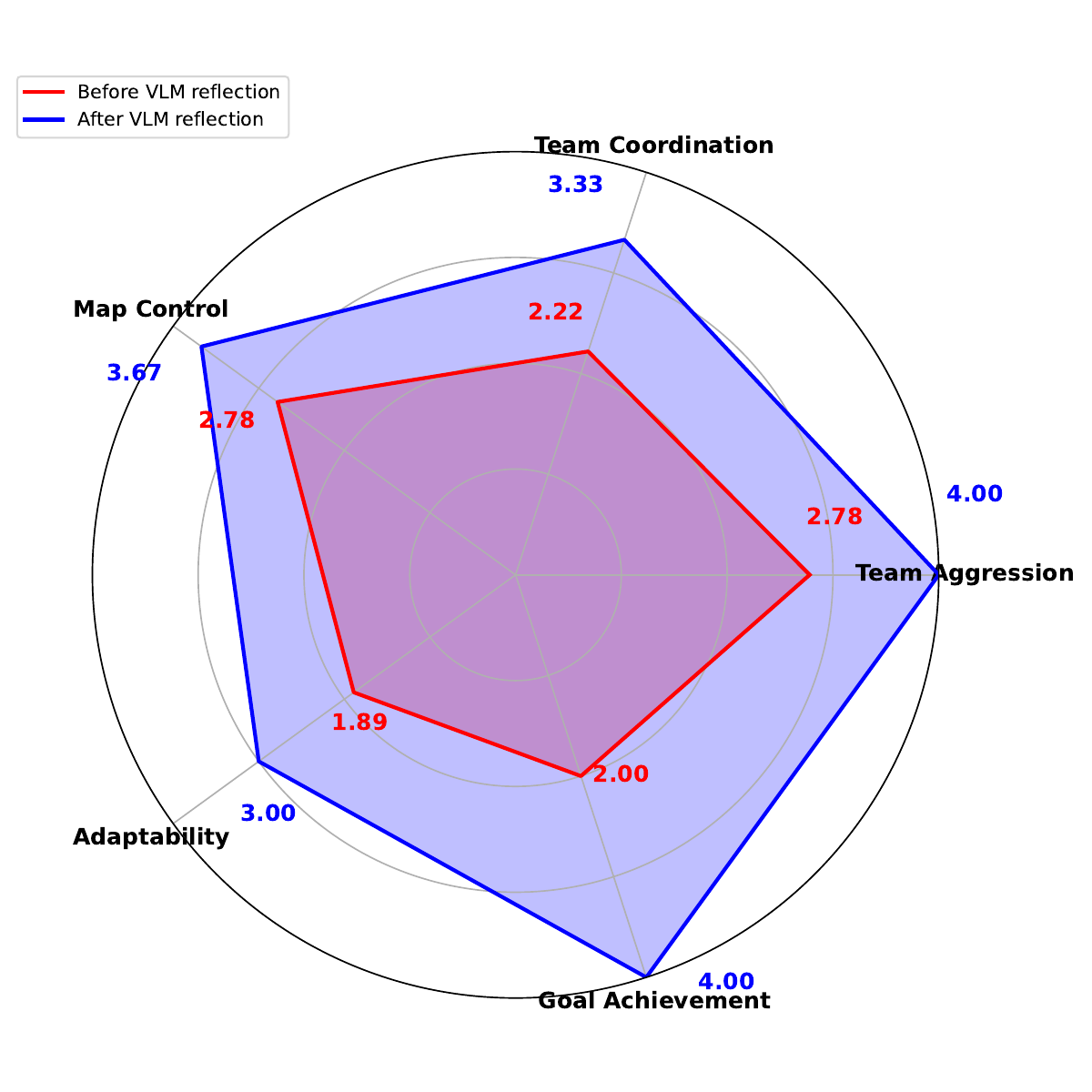}
    \caption{Effects of VLM reflexion on game plays.}
    \label{fig:vlm_analysis}
\end{figure}

For each dimension, we employ specialized prompts that guide the VLM to provide detailed analysis with specific justifications and quantitative ratings. The resulting multi-dimensional assessment, visualized in Figure~\ref{fig:vlm_analysis}, provides a comprehensive view of tactical performance that complements traditional numerical metrics.
The integration of VLM-based analysis into our Reflexion module has demonstrated significant benefits for behavior tree generation. As illustrated in Figure~\ref{fig:vlm_analysis}, policy improvements guided by VLM analysis show consistent enhancement across all five tactical dimensions. This improvement pattern highlights the complementary nature of VLM-based reflection and traditional numerical assessment -- while numerical metrics effectively capture local performance indicators (e.g., accuracy, resource efficiency), VLM analysis excels at identifying global tactical patterns that might not be immediately apparent from statistical measures alone.

\subsection{Playing Thousands of First-Person Shooter (FPS) Games}

\begin{figure}[h!] % Use [h!] to encourage placement here, but adjust as needed
\centering % Center the whole figure environment

\begin{subfigure}{0.48\textwidth} % Adjust width as needed
  \centering
  \includegraphics[width=\linewidth]{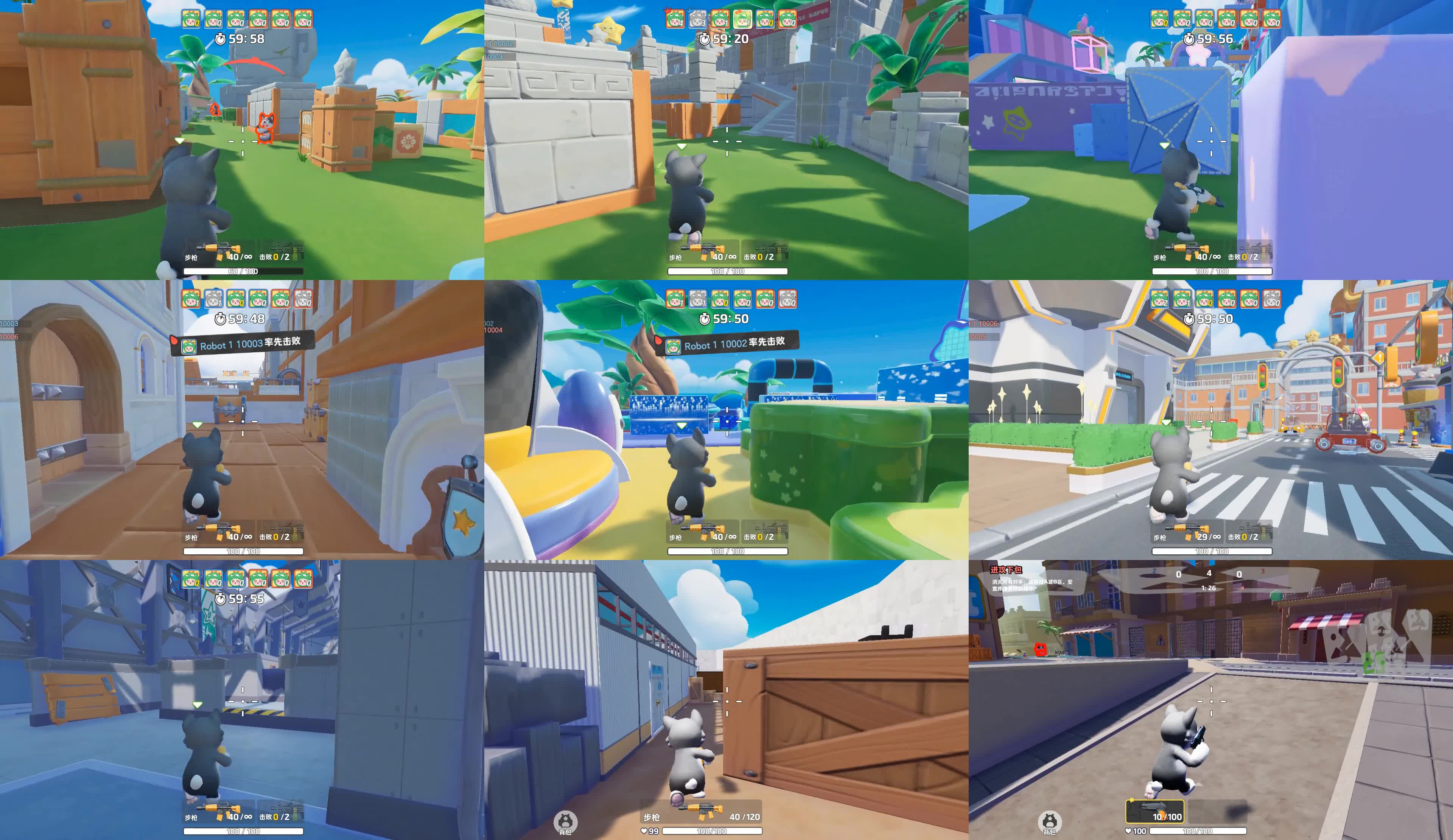} % Replace with your first image file
  %\caption{First subfigure}
  \label{fig:sub1}
\end{subfigure}\hfill % Key for horizontal spacing: \hfill pushes them apart
\begin{subfigure}{0.48\textwidth} % Adjust width as needed
  \centering
  \includegraphics[width=\linewidth]{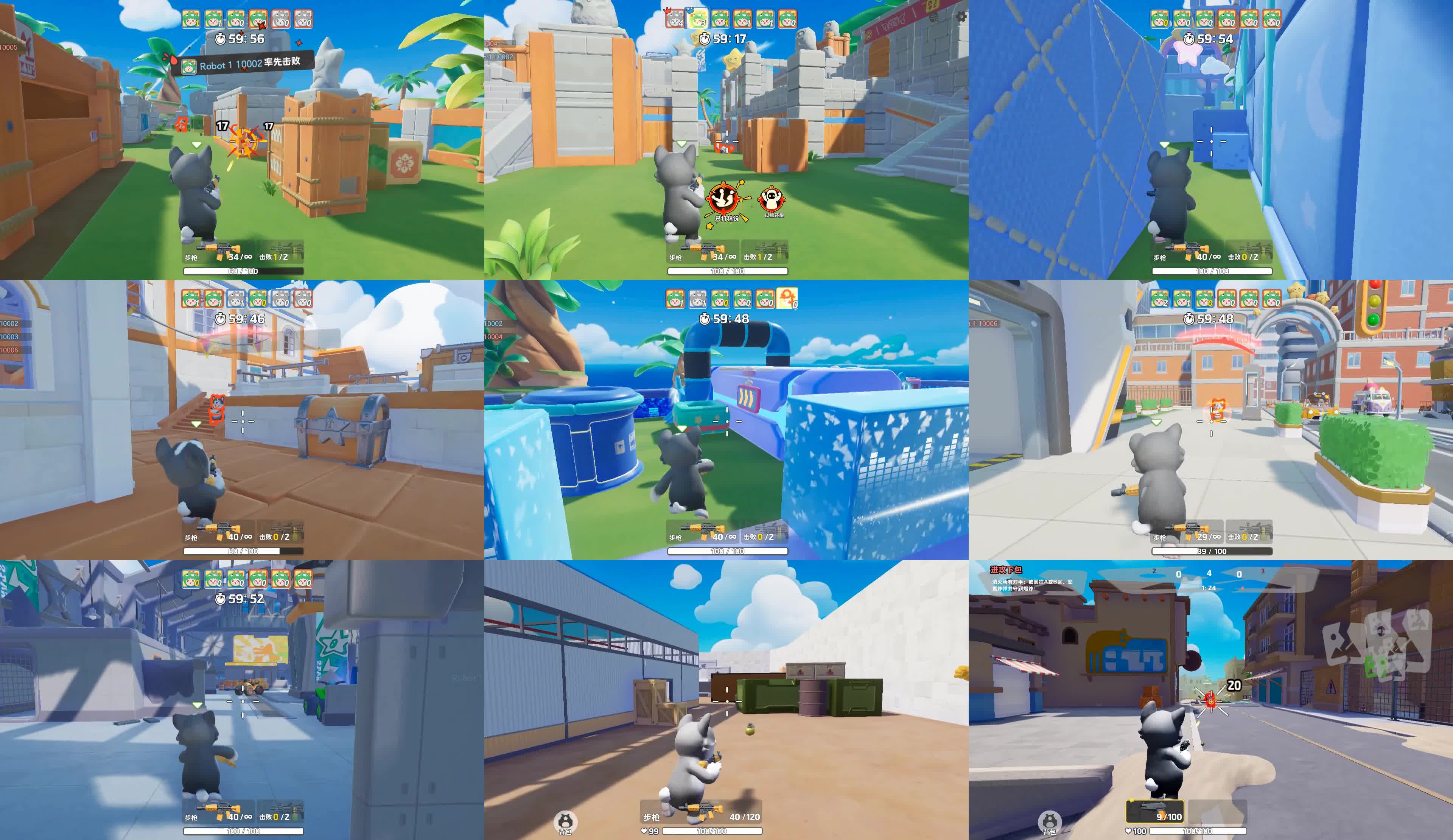} % Replace with your second image file
  %\caption{Second subfigure}
  \label{fig:sub2}
\end{subfigure}
\caption{Illustrations of a basic policy playing $3\times 3 =9$ FPS games.}
\label{fig:9fps_general_policy}
\end{figure}

Figure~\ref{fig:9fps_general_policy} presents visual evidence of our policies operating across a diverse set of 9 first-person shooter games (arranged in a $3\times3$ grid). Video demonstrations, all the materials about the behavior trees (including illustration, DSL, and JSON files) are available on our \href{https://zhongwen.one/projects/PORTAL}{project website}. These results demonstrate the remarkable generalization capabilities of our hybrid policies, which successfully transfer across games with varying visual styles, mechanical implementations, and environmental configurations.

\begin{figure}[h!]
\begin{subfigure}{0.48\textwidth} % Adjust width as needed
  \centering
  \includegraphics[width=\linewidth]{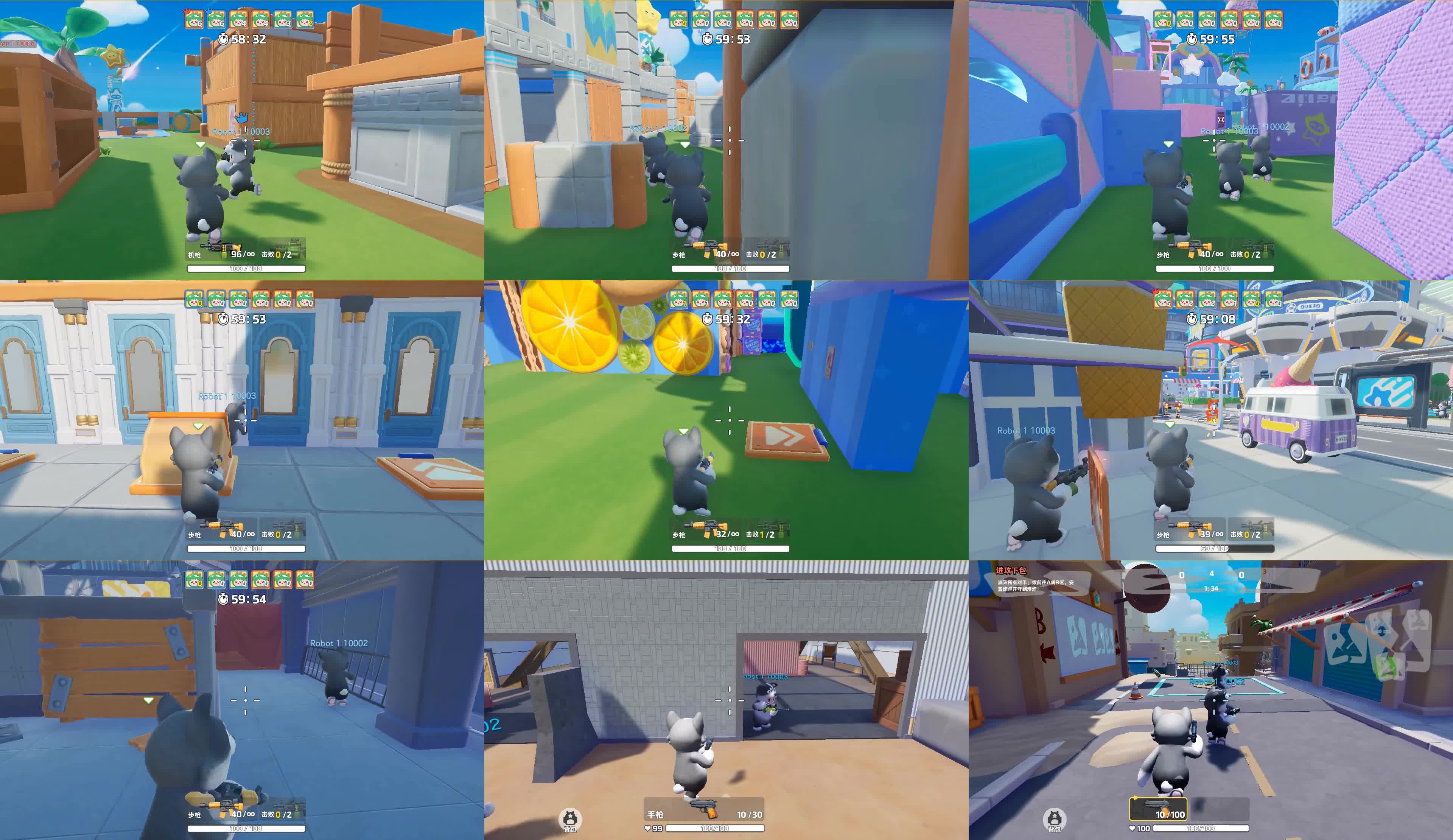} % Replace with your first image file
  %\caption{First subfigure}
  \label{fig:sub1}
\end{subfigure}\hfill % Key for horizontal spacing: \hfill pushes them apart
\begin{subfigure}{0.48\textwidth} % Adjust width as needed
  \centering
  \includegraphics[width=\linewidth]{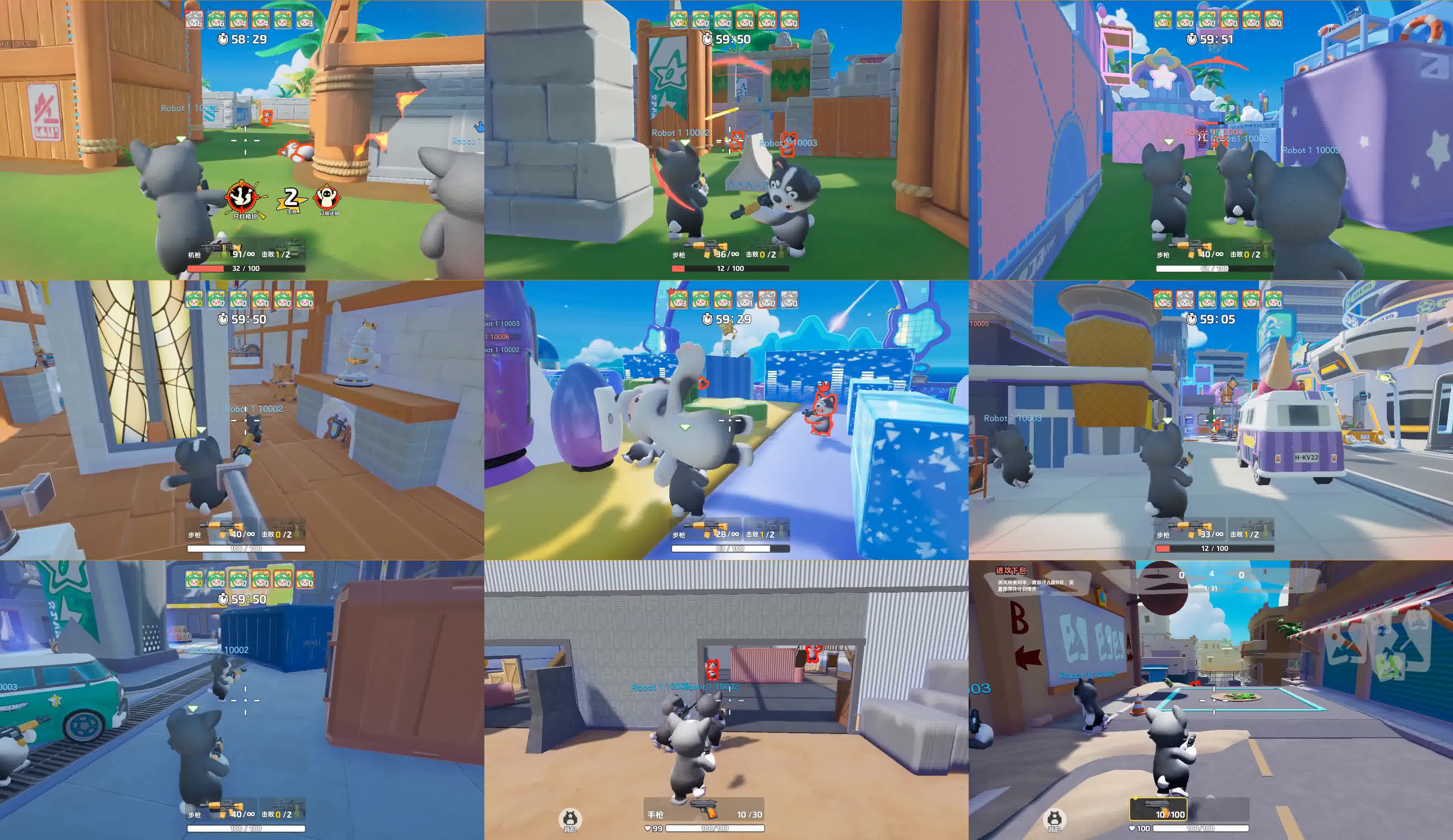} % Replace with your second image file
  %\caption{Second subfigure}
  \label{fig:sub2}
\end{subfigure}
\caption{Illustrations of a team attack policy playing $3\times 3 =9$ FPS games.}
\label{fig:9fps_team_attack}
\end{figure}

We present a case study focused on team coordination optimization. We applied our iterative refinement process with specific emphasis on team coordination metrics. The resulting progression, culminating in the structure shown in Figure~\ref{fig:9fps_team_attack}, demonstrates substantial qualitative improvements in collaborative behaviors, including coordinated positioning, covering fire patterns, and objective-focused role specialization. An additional example of advantage point policy is shown on our \href{https://zhongwen.one/projects/PORTAL}{project website}.

% \begin{figure}[h!]
% \begin{subfigure}{0.48\textwidth} % Adjust width as needed
%   \centering
%   \includegraphics[width=\linewidth]{PORTAL/figures/fps/advantage_point/video_0006.jpg} % Replace with your first image file
%   %\caption{First subfigure}
%   \label{fig:sub1}
% \end{subfigure}\hfill % Key for horizontal spacing: \hfill pushes them apart
% \begin{subfigure}{0.48\textwidth} % Adjust width as needed
%   \centering
%   \includegraphics[width=\linewidth]{PORTAL/figures/fps/advantage_point/video_0007.jpg} % Replace with your second image file
%   %\caption{Second subfigure}
%   \label{fig:sub2}
% \end{subfigure}
% \caption{Illustrations of an advantage-point policy playing $3\times 3 =9$ FPS games.}
% \label{fig:9fps_adv}
% \end{figure}

% Figure~\ref{fig:9fps_adv} demonstrates another critical aspect of our system's behavioral optimization capabilities: the autonomous identification and utilization of tactical advantage points. In this example, the AI progressively learns to leverage terrain features within the game map, positioning itself at elevated or strategically valuable locations to maximize line-of-sight advantages for combat engagement.

\subsection{Instant Development Procedure}
A particularly significant advantage of our approach is its enablement of a novel and highly efficient deployment pipeline. Traditional approaches to behavioral adjustment in game AI typically require computationally expensive retraining of neural networks, resulting in substantial development delays and resource requirements. In contrast, our methodology facilitates immediate adaptation through dynamic behavior tree generation.
PORTAL generates new Behavior Trees based on specified criteria, exporting the resulting policy as a JSON configuration file that can be directly ingested by the game server. This mechanism enables real-time adaptation of agent behavior within the game environment without requiring any recompilation or neural network retraining.
The practical implications of this capability are substantial:

\begin{itemize}
    \item \textbf{Rapid Iteration}: Game designers can specify behavioral modifications in natural language, receive an updated policy within minutes, and immediately test the results in-game.
    \item \textbf{Dynamic Difficulty Adjustment}: Servers can automatically generate customized opponent behaviors tailored to specific player skill levels or play styles.
    \item \textbf{Environment-Specific Adaptation}: Policies can be dynamically adjusted to accommodate new game environments, objectives, or rule modifications without developer intervention.
\end{itemize}

This instant development procedure represents a paradigm shift in game AI deployment, transitioning from the traditional cycle of specification, implementation, training, and deployment to a streamlined process where natural language descriptions directly translate to executable policies. The resulting efficiency gains not only accelerate development but also enable entirely new approaches to dynamic content generation and personalized gaming experiences.
\section{Discussions}\label{sec:dis}
% \textbf{Development Efficiency}: We have demonstrated how our designed system can streamline the development of game-playing AIs to enable agents that play thousands of 3D video games. Compared to classical reinforcement learning with massive simulation and distributed training, our development efficiency is accelerated by a factor of approximately \todo{500x}. Traditional reinforcement learning approaches often require days or weeks of training, while GenBT can generate, test, and refine a behavior tree within minutes, representing a significant advancement in AI development efficiency for complex game environments.

\textbf{Beyond FPS Games}: The methodology described in this paper extends readily beyond the first-person shooter genre explored in our experiments. The core architecture -- leveraging LLMs to generate behavior trees expressed in DSL -- provides a flexible framework applicable to diverse game genres and interaction paradigms.
Adapting our approach to new game categories follows a standardized procedure:

\begin{itemize}
    \item \textbf{Identify the basic action nodes} required for the target genre, such as resource management in strategy games, spell casting in RPGs, or evasive maneuvers in racing games.
    \item \textbf{Train focused neural networks} to handle specific navigation and skill requirements for the targeted genre, ensuring effective execution of low-level tasks.
    \item \textbf{Modify the prompting} framework to incorporate relevant game mechanics descriptions and desired tactical strategies appropriate for the new context.
\end{itemize}

The fundamental decomposition principle -- separating high-level strategic planning from low-level execution -- remains consistent across genres, enabling rapid adaptation to new gaming paradigms. This versatility stems from the generality of the behavior tree formalism and the flexibility of our DSL representation, which can accommodate diverse action spaces and strategic considerations without structural changes to the underlying architecture.

\begin{figure}[htp]
\centering
\includegraphics[width=0.3\linewidth]{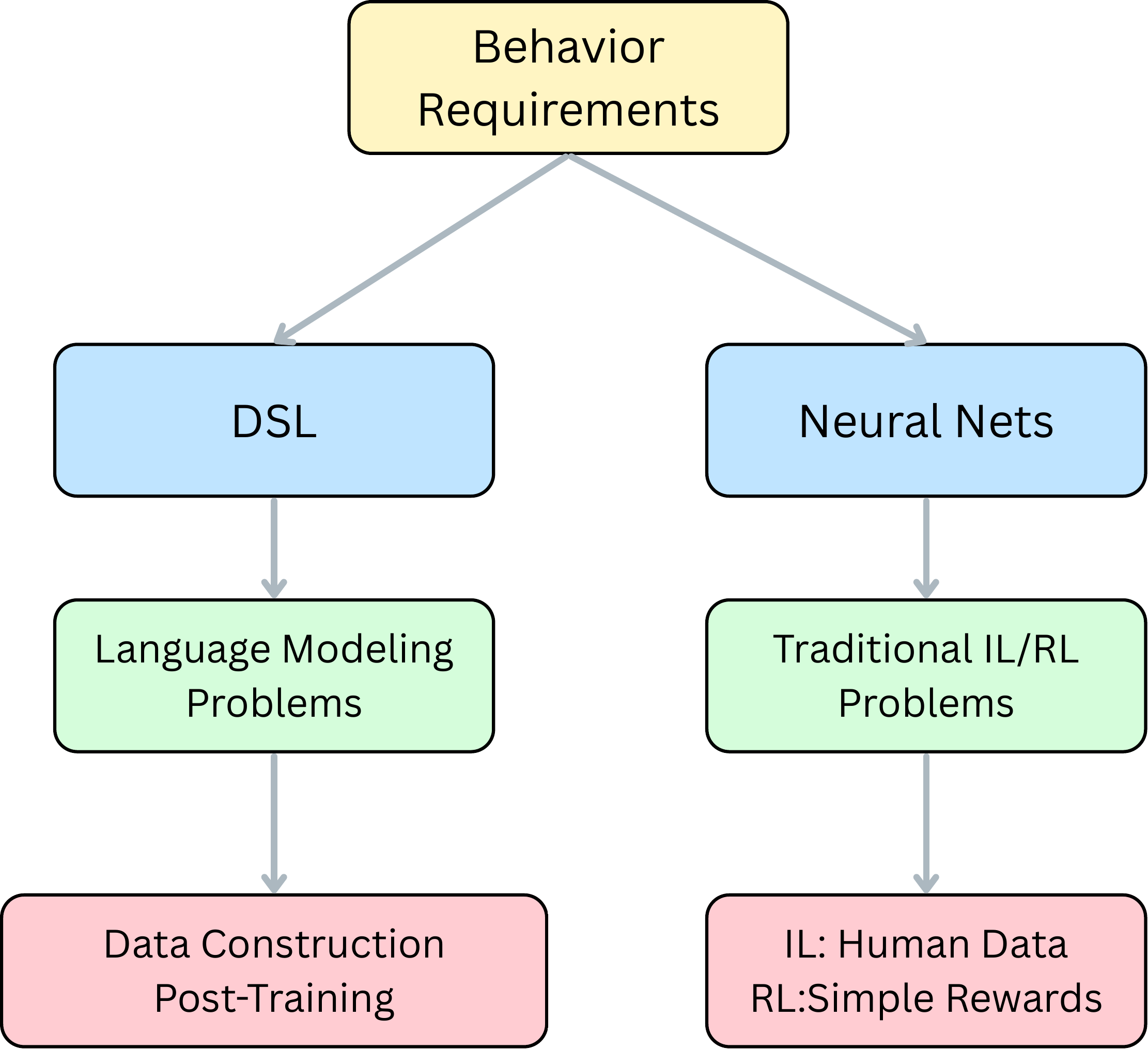}
\caption{Decomposition of AI behavior requirements and How to tackle them.}
\label{fig:behavior_reqs}
\end{figure}

\textbf{Generalization}: A persistent challenge in traditional reinforcement learning is achieving robust generalization to unseen environments. Conventional RL agents typically exhibit substantial performance degradation when confronted with even minor variations in environment dynamics or reward structures—a phenomenon attributable to overfitting to specific training conditions.
Our approach addresses this challenge through a two -- tiered generalization strategy as illustrated in Figure~\ref{fig:behavior_reqs}:

\begin{itemize}
    \item \textbf{Tactical-Level Generalization}: The high-level strategic structure, represented through behavior trees or DSL expressions, captures generalizable decision principles that transfer effectively across environments. For example, the fundamental strategic patterns in FPS games -- such as prioritizing threat elimination, position optimization, and resource management -- remain consistent regardless of specific game implementation details. These tactical structures transfer seamlessly between games from different studios or created by different developers, providing a robust foundation for cross-environment performance.
    \item \textbf{Action-Level Generalization}: At the execution level, we parameterize specific actions using focused neural networks trained with simplified, single-dimensional reward functions. This decomposition represents a significant departure from conventional approaches that attempt to optimize multi-objective reward functions simultaneously. By isolating specific objectives, such as navigation efficiency, target acquisition, or threat avoidance, we create more generalizable task-specific policies that perform consistently across diverse environmental conditions.
\end{itemize}

The decomposition inherent in our hybrid architecture converts complex, multi-objective optimization problems into collections of simpler, more tractable subproblems. This approach mitigates the ambiguity and conflicting objectives often present in monolithic reward functions, resulting in substantially improved generalization. Each neural component focuses on a well-defined task with clear success criteria, enabling more effective learning and transfer across environments.

\textbf{Diversity of Behaviors}: Behavioral diversity represents a critical factor in creating engaging and realistic game environments populated by non-player characters. Our framework facilitates diverse agent behaviors through systematic variation at both tactical and execution levels as shown in Figure~\ref{fig:behavior_reqs}.
At the tactical level, diversity emerges from several sources: variations in the prompting descriptions provided to the LLM, resulting in structurally distinct behavior trees;
differences in the available node sets accessible to the DSL, constraining or expanding strategic possibilities;
explicit diversity objectives incorporated into the generation process, encouraging exploration of alternative strategic approaches.

At the execution level, diversity manifests through: training local control networks on different distributions of human data, capturing varying skill levels and play styles; introducing controlled stochasticity into neural network outputs, creating natural variations in execution; parameterizing reward functions to emphasize different execution characteristics, such as aggression, caution, or efficiency.

The hierarchical decomposition of our approach makes achieving behavioral diversity substantially more tractable than in classical reinforcement learning systems. Rather than attempting to discover diverse behaviors through environmental exploration -- a computationally intensive process with limited guarantees, we can explicitly engineer diversity through structural variations in the generated policies. This approach provides game designers with unprecedented control over the distribution and characteristics of AI behaviors, enabling the creation of rich, varied game experiences with minimal development overhead.

\textbf{Multi-agent Collaboration}: While our primary focus has been on single-agent scenarios, our framework extends naturally to multi-agent collaboration contexts. We adapt our approach to collaborative settings by drawing inspiration from the Centralized Training with Decentralized Execution (CTDE) paradigm prevalent in multi-agent reinforcement learning.
For multi-agent scenarios, we extend our DSL to incorporate joint actions and states involving multiple agents. In this extended formalism, nodes can specify coordinated behaviors such as "agents A and B move to flanking positions while agent C provides covering fire." This higher-level representation captures the strategic coordination essential for effective team play.
Following the construction of this multi-agent DSL representation, we decompose it into agent-specific behavior trees, enabling decentralized execution. Each agent receives an individualized behavior tree derived from the joint strategy, tailored to its specific role within the coordinated plan. This decomposition preserves the strategic coordination specified in the joint representation while enabling autonomous execution by individual agents. The resulting multi-agent system exhibits coordinated behaviors emerging from explicit strategic planning rather than implicit policy convergence, yielding more interpretable and controllable collaborative dynamics than typically achieved through end-to-end multi-agent reinforcement learning.
% This approach offers several advantages for multi-agent scenarios:

% \begin{itemize}
%     \item Explicit modeling of coordination strategies at the DSL level, making collaborative intentions transparent and modifiable
%     \item Decentralized execution that eliminates the need for continuous communication between agents during gameplay
%     \item Role specialization that emerges naturally from the decomposition process
%     \item Scalability to varying team sizes without fundamental architectural changes
% \end{itemize}

\textbf{Connections to Meta Chain-of-Thought}: Recent work on Meta Chain-of-Thought (Meta-CoT)~\cite{meta_cot} surveys techniques for enabling Large Language Models (LLMs) to perform complex reasoning, proposing a framework to model the reasoning process as:

$$p(\mathbf{a}, \mathbf{z}_1, \ldots, \mathbf{z}_n|\mathbf{x}) \propto \int \underbrace{p(\mathbf{a}, \mathbf{z}_1, \ldots, \mathbf{z}_n|\mathbf{z}_1, \ldots, \mathbf{z}_k, \mathbf{x})}_{\text{Joint Answer+CoT}} \underbrace{\prod_{t=1}^K p(\mathbf{z}_t | \mathbf{z}_{< t},\mathbf{x})}_{\text{Meta-CoT}} d\mathbf{z},
$$

where $\mathbf{x}$ represents the input, $\mathbf{a}$ is the final answer, and $\mathbf{z}_i$ are the intermediate reasoning steps (Chain-of-Thought). The process $\mathbf{x} \to \mathbf{z}_1 \to \ldots \to \mathbf{z}_K$ is referred to as Meta-CoT, representing a higher-level reasoning process about the \emph{reasoning steps themselves}. This framework emphasizes the iterative refinement of thought processes, a key aspect of System 2 reasoning.

Our methodology embodies a similar meta-reasoning paradigm but operates at a different level of abstraction. Rather than using LLMs to map environment states directly to actions, our system employs them to generate policy structures expressed in DSL. This distinction can be formalized as:

\begin{align*}
  \text{Policy: } &\pi: S \rightarrow A \\
  \text{Meta-Policy: } &\Pi: \text{Policy Description } \rightarrow \pi
\end{align*}

In this formulation, the meta-policy $\Pi$ accepts a description of a policy as input and produces an executable policy $\pi$ as output. This represents a meta-reasoning process where the LLM designs the policy structure rather than directly controlling agent actions. 

%This abstraction provides several advantages:

% \begin{itemize}
%     \item Enhanced Flexibility: By operating at the policy design level, the LLM can reason about strategic structures holistically rather than making isolated action decisions.
%     \item Constraint Integration: The policy design process can explicitly incorporate domain constraints, prior knowledge, and strategic principles without relying on them being implicitly discovered through interaction.
%     \item Computational Efficiency: The meta-reasoning process occurs offline, eliminating the latency associated with online LLM inference during gameplay.
% \end{itemize}

This approach positions LLMs as \emph{policy architects} rather than direct controllers, leveraging their reasoning capabilities for strategic design while delegating execution to more efficient specialized components. The parallels with Meta-CoT suggest promising directions for further research, particularly in developing more sophisticated meta-reasoning frameworks for policy generation and refinement.

\textbf{Tool Uses and Function Calls}: There are connections to tool uses~\cite{qin2023toolllm,schick2023toolformer} and function calls, but our proposed system offers distinct advantages. Tool uses enable language models to interact with external tools to overcome inherent limitations in performing specialized tasks. While one might conceptualize our proposed system as a form of tool use if neural network task nodes are viewed as tools, our approach fundamentally differs in several key aspects. The primary distinction is that previous work positions LLMs as actors interacting directly with the environment, whereas our system deploys LLMs as architects. In our framework, LLMs do not process environment states as inputs to determine tool or function selection, which enables our system to integrate seamlessly with real-time game engines. Importantly, all planning and Domain-Specific Language (DSL) architecture is executed offline rather than online, resulting in significant performance benefits for time-sensitive applications while maintaining the strategic advantages of LLM-based planning.

\textbf{Extensions to Other Applications}: The structural principles underlying our approach extend naturally beyond gaming to various other domains involving hierarchical control problems. The behavior tree DSL abstraction we have developed provides a generalizable framework for complex decision-making across diverse applications.
Particularly promising extension domains include:

\begin{itemize}
    \item \textbf{Embodied AI}: Robotic systems face similar challenges in decomposing complex tasks into manageable components while maintaining strategic coherence. Our approach offers a framework for generating interpretable policies that combine high-level strategic planning with specialized execution components tailored to specific robotic embodiments. An idea in the same principle has been explored in a concurrent work Gemini Robotics~\cite{geminirobotics2025}.
    \item \textbf{Autonomous Driving}: Vehicle control involves complex decision hierarchies spanning strategic route planning, tactical maneuver selection, and precise motion control. The DSL representation can capture these hierarchical relationships while neural components handle execution under varying environmental conditions.
\end{itemize}

Our approach offers two key advantages for these extended applications. First, the high-level policies expressed in DSL exhibit strong transfer capabilities across different tasks and environments, enabling knowledge reuse across similar problem domains. Second, the hierarchical decomposition inherent in the behavior tree structure significantly reduces the complexity of training complete end-to-end systems, facilitating better generalization across diverse embodiments and tasks.
Furthermore, the interpretability of the DSL representation addresses a critical requirement in many real-world applications, particularly those subject to safety constraints or regulatory oversight. By making policy structures explicit and readable, our approach facilitates analysis, verification, and targeted modification -- essential capabilities for deployment in sensitive domains.
\begin{section}{Conclusions}\label{sec:conclusion}
In this paper, we have presented PORTAL, a novel framework for generating game-playing AI agents using large language models to produce behavior trees expressed in domain-specific language. Our approach fundamentally reconceptualizes AI agent development for video games by transforming complex decision-making problems into language modeling tasks, enabling rapid policy generation and deployment across thousands of diverse gaming environments.
% The key innovations of our work include: \todo{rewrite}

% \begin{itemize}
%     \item A hybrid policy architecture that combines LLM-generated strategic structures with neural network task nodes, balancing interpretability with adaptive execution
%     \item A domain-specific language representation for behavior trees that facilitates generation by language models while maintaining precise execution semantics
%     \item A dual-feedback mechanism incorporating both quantitative metrics and vision-language model analysis to guide policy refinement
%     \item A level-by-level generation and reflexion process that enables targeted improvements to specific policy components
%     \item A policy scheduling network that dynamically selects appropriate behavior trees based on environmental conditions
% \end{itemize}

These innovations collectively address critical limitations of previous approaches. Unlike traditional reinforcement learning methods, our framework eliminates the need for massive simulation resources and distributed training, reducing development time from weeks or months to hours or minutes. In contrast to existing LLM-based approaches, our method avoids the latency issues associated with online inference, enabling deployment in games with strict real-time requirements.

Our experimental results demonstrate the effectiveness of this approach across thousands of first-person shooter games, showcasing significant improvements in development efficiency and policy generalization. The instant development procedure enabled by our framework represents a paradigm shift in game AI deployment, offering unprecedented capabilities for rapid iteration and adaptation.

PORTAL represents a significant step toward more efficient, adaptable, and capable artificial intelligence for interactive systems. By bridging the gap between the strategic reasoning capabilities of language models and the real-time performance requirements of dynamic environments, our approach opens new possibilities for the next generation of decision-making AI and beyond.
\end{section}

\bibliography{main}

\begin{thebibliography}{32}
\providecommand{\natexlab}[1]{#1}
\providecommand{\url}[1]{\texttt{#1}}
\expandafter\ifx\csname urlstyle\endcsname\relax
  \providecommand{\doi}[1]{doi: #1}\else
  \providecommand{\doi}{doi: \begingroup \urlstyle{rm}\Url}\fi

\bibitem[Anthropic(2025)]{claude3_7}
Anthropic.
\newblock Visible extended thinking in large language models.
\newblock \url{https://www.anthropic.com/research/visible-extended-thinking}, 2025.
\newblock Accessed: March 2025.

\bibitem[Berner et~al.(2019)Berner, Brockman, Chan, Cheung, Dębiak, Dennison, Farhi, Fischer, Hashme, Hunn, Luengo, Rae, H, McGrew, Nekoul, Plappert, Amodei, and Zaremba]{Dota2}
Christopher Berner, Greg Brockman, Brooke Chan, Vicki Cheung, Przemysław Dębiak, Christy Dennison, David Farhi, Quirin Fischer, Shariq Hashme, Chelsea Hunn, Imanol Luengo, Jack Rae, Rachel H, Bob McGrew, Tyna~Eloundou Nekoul, Matthias Plappert, Dario Amodei, and Wojciech Zaremba.
\newblock Dota 2 with large scale deep reinforcement learning.
\newblock \emph{arXiv preprint arXiv:1912.06680}, 2019.

\bibitem[Colledanchise and {\"O}gren(2018)]{BT}
Michele Colledanchise and Petter {\"O}gren.
\newblock \emph{Behavior Trees in Robotics and AI: An Introduction}.
\newblock CRC Press, 2018.

\bibitem[DeepSeek-AI et~al.(2025)DeepSeek-AI, Guo, Yang, Zhang, Song, Zhang, Xu, Zhu, Ma, Wang, Bi, Zhang, and (add other authors as needed or use~et al.)]{R1}
DeepSeek-AI, Daya Guo, Dejian Yang, Haowei Zhang, Junxiao Song, Ruoyu Zhang, Runxin Xu, Qihao Zhu, Shirong Ma, Peiyi Wang, Xiao Bi, Xiaokang Zhang, and ... (add other authors as needed or use~et al.).
\newblock {DeepSeek}-{R1}: Incentivizing reasoning capability in {LLMs} via reinforcement learning.
\newblock \emph{arXiv preprint arXiv:2501.12948}, 2025.

\bibitem[Espeholt et~al.(2018)Espeholt, Soyer, Munos, Simonyan, Mnih, Ward, Doron, Firoiu, Harley, Dunning, Legg, and Kavukcuoglu]{IMPALA}
Lasse Espeholt, Hubert Soyer, R{\'e}mi Munos, Karen Simonyan, Volodymyr Mnih, Tom Ward, Yotam Doron, Vlad Firoiu, Tim Harley, Iain Dunning, Shane Legg, and Koray Kavukcuoglu.
\newblock {IMPALA}: Scalable distributed deep-{RL} with importance weighted actor-learner architectures.
\newblock \emph{arXiv preprint arXiv:1802.01561}, 2018.

\bibitem[Gallotta et~al.(2024)Gallotta, Todd, Zammit, Earle, Liapis, Togelius, and Yannakakis]{llm_games}
Roberto Gallotta, Graham Todd, Marvin Zammit, Sam Earle, Antonios Liapis, Julian Togelius, and Georgios~N Yannakakis.
\newblock Large language models and games: A survey and roadmap.
\newblock \emph{IEEE Transactions on Games}, 2024.

\bibitem[{Gemini Robotics Team, Google DeepMind}(2025)]{geminirobotics2025}
{Gemini Robotics Team, Google DeepMind}.
\newblock Gemini {Robotics}: Bringing {AI} into the physical world.
\newblock \emph{arXiv preprint}, 2025.

\bibitem[Ma et~al.(2024)Ma, Mi, Zeng, Yan, Lin, Wu, Wang, and Zhang]{llm_sc2}
Weiyu Ma, Qirui Mi, Yongcheng Zeng, Xue Yan, Runji Lin, Yuqiao Wu, Jun Wang, and Haifeng Zhang.
\newblock Large language models play {StarCraft II}: Benchmarks and a chain of summarization approach.
\newblock \emph{Advances in Neural Information Processing Systems}, 37:\penalty0 133386--133442, 2024.

\bibitem[Mnih et~al.(2015)Mnih, Kavukcuoglu, Silver, Rusu, Veness, Bellemare, Graves, Riedmiller, Fidjeland, Ostrovski, Petersen, Beattie, Sadik, Antonoglou, King, Kumaran, Wierstra, Legg, and Hassabis]{DQN}
Volodymyr Mnih, Koray Kavukcuoglu, David Silver, Andrei~A. Rusu, Joel Veness, Marc~G. Bellemare, Alex Graves, Martin Riedmiller, Andreas~K. Fidjeland, Georg Ostrovski, Stig Petersen, Charles Beattie, Amir Sadik, Ioannis Antonoglou, Helen King, Dharshan Kumaran, Daan Wierstra, Shane Legg, and Demis Hassabis.
\newblock Human-level control through deep reinforcement learning.
\newblock \emph{Nature}, 518\penalty0 (7540):\penalty0 529--533, 2015.
\newblock \doi{10.1038/nature14236}.

\bibitem[OpenAI et~al.(2023)OpenAI, Achiam, Adler, Agarwal, Ahmad, Akkaya, Aleman, Almeida, Altman, Alvarez, Anderson, Anderson, Aneja, Anton, Askell, Aslam, Azer, Bach, Bai, Balwit, Banks, Batmanghelich, Baxter, Beres, Biderman, Burnell, Achille, et~al.]{GPT4}
OpenAI, Josh Achiam, Sasha Adler, Sandhini Agarwal, Sandeep Ahmad, Ilge Akkaya, Alty Aleman, David Almeida, Elie Altman, Alekh Alvarez, Haiming Anderson, Mira Anderson, Jyoti Aneja, Anthony Anton, Amanda Askell, Haidar Aslam, Alex Azer, Karsen Bach, Yuntao Bai, Mark Balwit, Kendra Banks, Kaylee Batmanghelich, Daniel Baxter, Pierre Beres, Stella Biderman, Nicholas Burnell, Alessandro Achille, et~al.
\newblock {GPT-4} technical report.
\newblock \emph{arXiv preprint arXiv:2303.08774}, 2023.

\bibitem[PrismarineJS(2025)]{mineflayer}
PrismarineJS.
\newblock {Mineflayer}: A high-level {JavaScript} {API} for creating {Minecraft} bots.
\newblock \url{https://github.com/PrismarineJS/mineflayer}, 2025.
\newblock Accessed: 2025-03-06.

\bibitem[Qin et~al.(2023)Qin, Liang, Ye, Liu, Zhang, Han, Jin, Yang, Xue, Han, and Tang]{qin2023toolllm}
Yujia Qin, Shengding Liang, Houqiang Ye, Ge~Liu, Yi~Zhang, Weizhu Han, Xin Jin, Lifan Yang, Nianwen Xue, Jiawei Han, and Jie Tang.
\newblock {ToolLLM}: Facilitating large language models to master 16000+ real-world {APIs}.
\newblock \emph{arXiv preprint arXiv:2307.16789}, 2023.

\bibitem[Schick et~al.(2023)Schick, Dwivedi-Yu, Dess{\`i}, Raileanu, Lomeli, Zettlemoyer, Cancedda, and Scialom]{schick2023toolformer}
Timo Schick, Jane Dwivedi-Yu, Roberto Dess{\`i}, Roberta Raileanu, Maria Lomeli, Luke Zettlemoyer, Nicola Cancedda, and Thomas Scialom.
\newblock Toolformer: Language models can teach themselves to use tools.
\newblock \emph{arXiv preprint arXiv:2302.04761}, 2023.

\bibitem[Schulman et~al.(2017)Schulman, Wolski, Dhariwal, Radford, and Klimov]{PPO}
John Schulman, Filip Wolski, Prafulla Dhariwal, Alec Radford, and Oleg Klimov.
\newblock Proximal policy optimization algorithms.
\newblock \emph{arXiv preprint arXiv:1707.06347}, 2017.

\bibitem[Shannon(1950)]{shannon1950programming}
Claude~E. Shannon.
\newblock Programming a computer for playing chess.
\newblock \emph{The London, Edinburgh, and Dublin Philosophical Magazine and Journal of Science}, 41\penalty0 (314):\penalty0 256--275, 1950.
\newblock \doi{10.1080/14786445008521796}.

\bibitem[Shinn et~al.(2023)Shinn, Cassano, Berman, Gopinath, Narasimhan, and Yao]{reflexion}
Noah Shinn, Federico Cassano, Edward Berman, Ashwin Gopinath, Karthik Narasimhan, and Shunyu Yao.
\newblock Reflexion: Language agents with verbal reinforcement learning, 2023.

\bibitem[Silver et~al.(2016)Silver, Huang, Maddison, Guez, Sifre, van~den Driessche, Schrittwieser, Antonoglou, Panneershelvam, Lanctot, Dieleman, Grewe, Nham, Kalchbrenner, Sutskever, Lillicrap, Leach, Kavukcuoglu, Graepel, and Hassabis]{AlphaGo}
David Silver, Aja Huang, Chris~J. Maddison, Arthur Guez, Laurent Sifre, George van~den Driessche, Julian Schrittwieser, Ioannis Antonoglou, Veda Panneershelvam, Marc Lanctot, Sander Dieleman, Dominik Grewe, John Nham, Nal Kalchbrenner, Ilya Sutskever, Timothy Lillicrap, Madeleine Leach, Koray Kavukcuoglu, Thore Graepel, and Demis Hassabis.
\newblock Mastering the game of {Go} with deep neural networks and tree search.
\newblock \emph{Nature}, 529\penalty0 (7587):\penalty0 484--489, 2016.
\newblock \doi{10.1038/nature16961}.

\bibitem[Silver et~al.(2017)Silver, Schrittwieser, Simonyan, Antonoglou, Huang, Guez, Hubert, Baker, Lai, Bolton, Chen, Lillicrap, Hui, Sifre, van~den Driessche, Graepel, and Hassabis]{AlphaGoZero}
David Silver, Julian Schrittwieser, Karen Simonyan, Ioannis Antonoglou, Aja Huang, Arthur Guez, Thomas Hubert, Lucas Baker, Matthew Lai, Adrian Bolton, Yutian Chen, Timothy Lillicrap, Fan Hui, Laurent Sifre, George van~den Driessche, Thore Graepel, and Demis Hassabis.
\newblock Mastering the game of {Go} without human knowledge.
\newblock \emph{Nature}, 550\penalty0 (7676):\penalty0 354--359, 2017.
\newblock \doi{10.1038/nature24270}.

\bibitem[Silver et~al.(2018)Silver, Hubert, Schrittwieser, Antonoglou, Lai, Guez, Lanctot, Sifre, Kumaran, Graepel, Lillicrap, Simonyan, and Hassabis]{AlphaZero}
David Silver, Thomas Hubert, Julian Schrittwieser, Ioannis Antonoglou, Matthew Lai, Arthur Guez, Marc Lanctot, Laurent Sifre, Dharshan Kumaran, Thore Graepel, Timothy Lillicrap, Karen Simonyan, and Demis Hassabis.
\newblock A general reinforcement learning algorithm that masters {Chess}, {Shogi}, and {Go} through self-play.
\newblock \emph{Science}, 362\penalty0 (6419):\penalty0 1140--1144, 2018.
\newblock \doi{10.1126/science.aar6404}.

\bibitem[Sutton and Barto(2018)]{RL_book}
Richard~S Sutton and Andrew~G Barto.
\newblock \emph{Reinforcement Learning: An Introduction}.
\newblock MIT press, 2018.

\bibitem[Sutton et~al.(1999)Sutton, Precup, and Singh]{option}
Richard~S Sutton, Doina Precup, and Satinder Singh.
\newblock Between {MDPs} and semi-{MDP}s, a framework for temporal abstraction in reinforcement learning.
\newblock \emph{Artificial intelligence}, 112\penalty0 (1-2):\penalty0 181--211, 1999.

\bibitem[Team et~al.(2024)Team, Georgiev, Lei, Burnell, Bai, Gulati, Tanzer, Vincent, Pan, Wang, et~al.]{Gemini}
Gemini Team, Petko Georgiev, Ving~Ian Lei, Ryan Burnell, Libin Bai, Anmol Gulati, Garrett Tanzer, Damien Vincent, Zhufeng Pan, Shibo Wang, et~al.
\newblock Gemini 1.5: Unlocking multimodal understanding across millions of tokens of context.
\newblock \emph{arXiv preprint arXiv:2403.05530}, 2024.

\bibitem[Team(2024{\natexlab{a}})]{qwen25}
Qwen Team.
\newblock {Qwen2.5}: An innovative, generalist, and open large language model family.
\newblock \emph{arXiv preprint}, June 2024{\natexlab{a}}.
\newblock URL \url{https://qwenlm.github.io/blog/qwen2.5/}.
\newblock Alibaba Cloud.

\bibitem[Team(2024{\natexlab{b}})]{qwen25coder}
Qwen Team.
\newblock {Qwen2.5-Coder}: An enhanced large language model for code understanding and generation.
\newblock \emph{arXiv preprint}, June 2024{\natexlab{b}}.
\newblock URL \url{https://qwenlm.github.io/blog/qwen2.5-coder/}.
\newblock Alibaba Cloud.

\bibitem[Tsai et~al.(2023)Tsai, Zhou, Liu, Li, Yu, and Mei]{Zork}
Chen~Feng Tsai, Xiaochen Zhou, Sierra~S Liu, Jing Li, Mo~Yu, and Hongyuan Mei.
\newblock Can large language models play text games well? {Current} state-of-the-art and open questions.
\newblock \emph{arXiv preprint arXiv:2304.02868}, 2023.

\bibitem[Vinyals et~al.(2019)Vinyals, Babuschkin, Czarnecki, Mathieu, Dudzik, Chung, Choi, Powell, Ewalds, Eccles, Casares, Budden, Osindero, Veliantsev, Agapiou, Devlin, Tassa, Tracey, Hassabis, and Kavukcuoglu]{AlphaStar}
Oriol Vinyals, Igor Babuschkin, Wojciech~M. Czarnecki, Michaël Mathieu, Andrew Dudzik, Junyoung Chung, David~H. Choi, Richard Powell, Timo Ewalds, Tom Eccles, Norman Casares, Thomas Budden, Simon Osindero, Aliaksei Veliantsev, Johannes Agapiou, James Devlin, Yuval Tassa, Brendan Tracey, Demis Hassabis, and Koray Kavukcuoglu.
\newblock Grandmaster level in {StarCraft II} using multi-agent reinforcement learning.
\newblock \emph{Nature}, 575\penalty0 (7782):\penalty0 350--354, 2019.
\newblock \doi{10.1038/s41586-019-1724-z}.

\bibitem[Wang et~al.(2023)Wang, Xie, Jiang, Mandlekar, Xiao, Zhu, Fan, and Anandkumar]{Voyager}
Guanzhi Wang, Yuqi Xie, Yunfan Jiang, Ajay Mandlekar, Chaowei Xiao, Yuke Zhu, Linxi Fan, and Anima Anandkumar.
\newblock {VOYAGER}: An open-ended embodied agent with large language models.
\newblock \emph{Transactions on Machine Learning Research}, 2023.

\bibitem[Wang et~al.(2024)Wang, Li, Song, Xu, Tang, Zhuge, Pan, Song, Li, Singh, Tran, Li, Ma, Zheng, Qian, Shao, Muennighoff, Zhang, Hui, Lin, Brennan, Peng, Ji, and Neubig]{OpenHands}
Xingyao Wang, Boxuan Li, Yufan Song, Frank~F. Xu, Xiangru Tang, Mingchen Zhuge, Jiayi Pan, Yueqi Song, Bowen Li, Jaskirat Singh, Hoang~H. Tran, Fuqiang Li, Ren Ma, Mingzhang Zheng, Bill Qian, Yanjun Shao, Niklas Muennighoff, Yizhe Zhang, Binyuan Hui, Junyang Lin, Robert Brennan, Hao Peng, Heng Ji, and Graham Neubig.
\newblock {OpenHands: An Open Platform for AI Software Developers as Generalist Agents}, 2024.

\bibitem[Wei et~al.(2022)Wei, Wang, Schuurmans, Bosma, Ichter, Xia, Chi, Le, and Zhou]{CoT}
Jason Wei, Xuezhi Wang, Dale Schuurmans, Maarten Bosma, Brian Ichter, Fei Xia, Ed~Chi, Quoc Le, and Denny Zhou.
\newblock {Chain of Thought} prompting elicits reasoning in {Large Language Models}.
\newblock \emph{Advances in Neural Information Processing Systems}, 35:\penalty0 24824--24837, 2022.

\bibitem[Xiang et~al.(2025)Xiang, Snell, Gandhi, Albalak, Singh, Blagden, Phung, Rafailov, Lile, Mahan, et~al.]{meta_cot}
Violet Xiang, Charlie Snell, Kanishk Gandhi, Alon Albalak, Anikait Singh, Chase Blagden, Duy Phung, Rafael Rafailov, Nathan Lile, Dakota Mahan, et~al.
\newblock Towards {System 2} reasoning in {LLMs}: Learning how to think with meta {Chain-of-Thought}.
\newblock \emph{arXiv preprint arXiv:2501.04682}, 2025.

\bibitem[Yao et~al.(2023)Yao, Zhao, Yu, Du, Shafran, Narasimhan, and Cao]{ReAct}
Shunyu Yao, Jeffrey Zhao, Dian Yu, Nan Du, Izhak Shafran, Karthik Narasimhan, and Yuan Cao.
\newblock {ReAct}: Synergizing reasoning and acting in language models.
\newblock \emph{arXiv preprint arXiv:2210.03629}, 2023.

\bibitem[Ye et~al.(2020)Ye, Liu, Sun, Shi, Zhao, Wu, Yu, Yang, Wu, Guo, Chen, Yin, Zhang, Shi, Wang, Fu, Yang, and Huang]{wzry}
Deheng Ye, Zhao Liu, Mingfei Sun, Bei Shi, Peilin Zhao, Hao Wu, Hongsheng Yu, Shaojie Yang, Xipeng Wu, Qingwei Guo, Qiaobo Chen, Yinyuting Yin, Hao Zhang, Tengfei Shi, Liang Wang, Qiang Fu, Wei Yang, and Lanxiao Huang.
\newblock Mastering complex control in {MOBA} games with deep reinforcement learning.
\newblock In \emph{AAAI}, pages 6672--6679. {AAAI} Press, 2020.

\end{thebibliography}

\newpage
\appendix
\onecolumn

\section{Prompt Template}
We use a prompt template as in OpenHands~\cite{OpenHands} which is rendered by Jinja2 engine to enable flexible prompt management. The template is extended with Markdowns from other folders and files to form a concrete prompt for the LLM to process.

\BeginSol
\# Task\\
You are a diligent AI programmer working on behavior trees design for game AI.\\
You use domain specific language (DSL) to behavior tree implementation.\\
Your goal is to implement the DSL given a specified tactic.\\

\#\# Game Scenario\\
\{\{instructions.scenarios.cs\}\}\\

\#\# Available Nodes\\
You are also given a collection of existing basic nodes in the game.\\
\{\{ instructions.actions.selector \}\}\\
\{\{ instructions.actions.sequence \}\}\\
\{\{ instructions.actions.condition \}\}\\
\{\{ instructions.actions.param \}\}\\
\{\{ instructions.actions.action \}\}\\

\#\# Tactics\\
\{\{ state.tactics \}\}\\

\#\# DSL Format\\
\{\{ instructions.format.dsl\_syntax \}\}\\

\#\# Response\\
\{\{ instructions.format.dsl\_nlu \}\}\\

\{\% if history.dsl\_tree \%\}\\
\#\#\# History Format Errors\\
Here's your last DSL implementation and its corresponding format problems, you should improve upon it and fix the format problems. \\

\{\{ history.dsl\_tree \}\}\\
\{\{ history.message \}\}\\
\{\% endif \%\}\\
\EndSol

\end{document}